\pdfoutput=1

\documentclass[11pt]{article}

\usepackage[final]{acl}

\usepackage{times}
\usepackage{latexsym}

\usepackage[T1]{fontenc}

\usepackage[utf8]{inputenc}

\usepackage{microtype}

\usepackage{inconsolata}

\usepackage{graphicx}

\usepackage{amsmath,graphicx}
\usepackage{booktabs} 
\usepackage{subcaption}
\usepackage{xcolor, colortbl}
\usepackage{algorithm}
\usepackage{algpseudocode}
\usepackage{xspace}
\usepackage{multirow}
\usepackage{xurl}
\usepackage{tikz}
\usepackage{amssymb}
\usepackage{pifont}
\newcommand{\cmark}{\ding{51}}%
\newcommand{\xmark}{\ding{55}}%
\usepackage{amsfonts}
\usepackage{mathtools, nccmath}

\usepackage{scalerel,xparse}
\NewDocumentCommand\emojifirst{}{
    \scalerel*{
        \includegraphics{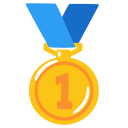}
    }{X}
}
\NewDocumentCommand\emojisecond{}{
    \scalerel*{
        \includegraphics{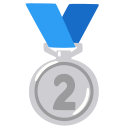}
    }{X}
}
\NewDocumentCommand\emojithird{}{
    \scalerel*{
        \includegraphics{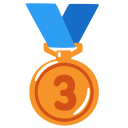}
    }{X}
}
\NewDocumentCommand\emojinone{}{
    \scalerel*{
        \includegraphics{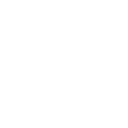}
    }{X}
}

\title{Towards Robust Speech Representation \\ Learning for Thousands of Languages}

\author{William Chen$^{1}$, Wangyou Zhang$^{1,2}$, Yifan Peng$^{1}$ Xinjian Li$^{1}$\\
        \textbf{Jinchuan Tian$^{1}$, Jiatong Shi$^{1}$, Xuankai Chang$^{1}$, Soumi Maiti$^{1}$} \\
        \textbf{Karen Livescu$^{1,3}$, Shinji Watanabe$^{1}$} \\
    $^{1}$ Carnegie Mellon University, $^{2}$ Shanghai Jiaotong University\\
    $^{3}$ Toyota Technological Institute at Chicago\\
  \texttt{williamchen@cmu.edu} \\}

\begin{document}
\maketitle
\begin{abstract}
Self-supervised learning (SSL) has helped extend speech technologies to more languages by reducing the need for labeled data. However, models are still far from supporting the world's 7000+ languages. We propose XEUS, a Cross-lingual Encoder for Universal Speech, trained on over 1 million hours of data across 4057 languages, extending the language coverage of SSL models 4-fold. We combine 1 million hours of speech from existing publicly accessible corpora with a newly created corpus of 7400+ hours from 4057 languages, which will be publicly released. To handle the diverse conditions of multilingual speech data, we augment the typical SSL masked prediction approach with a novel dereverberation objective, increasing robustness. We evaluate XEUS on several benchmarks, and show that it consistently outperforms or achieves comparable results to state-of-the-art (SOTA) SSL models across a variety of tasks. XEUS sets a new SOTA on the ML-SUPERB benchmark: it outperforms MMS 1B and w2v-BERT 2.0 v2 by 0.8\% and 4.4\% respectively, despite having less parameters or pre-training data. Checkpoints, code, and data are found in \url{https://www.wavlab.org/activities/2024/xeus/}.
\end{abstract}

\vspace{-0.3cm}
\section{Introduction}
\vspace{-0.1cm}
Our planet is home to over 7000 languages \cite{Austin2011-AUSTCH-2}, yet most speech processing models are only capable of serving at most 100-150 of them \cite{barrault2023seamless, whisper}. The biggest constraint in supporting more of languages is the lack of transcribed speech: only around half of the world's languages have a formal writing system \cite{ethnologue}, and even fewer of them have the resources to support the scale of annotated data required for training neural models. A common approach to address this limitation is self-supervised learning (SSL) on large amounts of unlabeled multilingual speech \cite{google-usm, wilderness, asr2k}, which allows for strong downstream performance even when access to annotated data is limited. 

While SSL models have more relaxed data requirements relative to supervised models, few works have fully exploited this aspect to scale models to more languages. In fact, most multilingual SSL models remain in the 50-150 language range of coverage \cite{babu2021xls, wavelablm, chiu2022self}, reducing the benefits of this advantage. The MMS project \cite{pratap2023scaling} sought to address this issue by directly crawling data for more languages, scaling SSL pre-training to 1,000+ languages. The authors collected speech across 1,406 languages and used it to train an SSL model, showing state-of-the-art (SOTA) results after fine-tuning on multilingual automatic speech recognition (ASR) and 3,900-way language identification (LID). While the MMS models were publicly released, the crawled data was not, preventing it from being used in future work and thus the models from being reproduced (Table \ref{tab:models}).

An additional issue that is relatively unexplored in SSL research is robustness to noisy data. This aspect is important for multilingual models, since the available recordings of low-resource languages tend to be particularly noisy \cite{commonvoice}. The issue is exacerbated by the fact that existing multilingual SSL corpora lack diversity not only in languages but also in speaking style and recording conditions. WavLM \cite{ChenWavLm} and WavLabLM \cite{wavelablm} tackled this issue by simulating noisy conditions during training, overlapping utterances or adding acoustic noise to simulate multi-speaker and noisy environments respectively. While effective, we believe that this technique can be improved to cover an even wider variety of recording conditions.

Our goal is to thus build a universal speech encoder that can handle both linguistically and acoustically diverse speech. To achieve this, we propose XEUS (pronounced Zeus) --- a Cross-lingual Encoder for Universal Speech. XEUS is an E-Branchformer \cite{ebf} encoder pre-trained on over 1 million hours of publicly available data across a wide variety of recording conditions. We first curate the data from 37 existing corpora to ensure a diverse selection of speech and recording conditions not often found in standard ASR datasets, including but not limited to spontaneous speech, accented speech, code-switching, indigenous languages, and singing voices. We expand the language coverage of XEUS by introducing a new SSL corpus that uses data sources previously unseen in speech literature. This corpus, which will be publicly released, contains 7,413 hours of unlabeled audio across 4,057 languages, the widest coverage of any speech processing dataset. 

To enhance the model's robustness, XEUS is also pre-trained with a novel SSL objective of acoustic dereverberation, which requires the model to predict clean discrete phonetic pseudo-labels from simulated reverberant audio. By combining this objective with HuBERT-style masked prediction \cite{hsuHubert} and WavLM-style denoising \cite{ChenWavLm}, XEUS is designed to be the next step towards a truly universal speech encoder for any language or recording condition. 

In our downstream evaluations, we find that XEUS consistently improves over SOTA SSL models across a wide variety of tasks. XEUS sets a new SOTA on the ML-SUPERB multilingual ASR benchmark, outperforming SSL models such as MMS \cite{pratap2023scaling} and w2v-BERT 2.0 v2 \cite{barrault2023seamless} while having fewer parameters or less training data. Our speech translation (ST) results show the effectiveness of XEUS' wide language coverage, even for languages with less than 10 hours of data in the pre-training corpus. We also explore XEUS' potential in generative tasks and show its superiority on speech resynthesis when compared to other SSL encoders. Finally, we evaluate XEUS' representations on a variety of tasks through the English-only SUPERB benchmark, where it sets a new SOTA on 4 tasks despite XEUS' focus on multilingual performance.

To conduct SSL pre-training at such scale, we had to make significant optimizations to existing speech processing toolkits. To encourage further SSL research and reproducibility, we will publicly release this code, along with the training configurations and checkpoints for XEUS. We also release all 200+ intermediate checkpoints and training logs obtained throughout the pre-training for further research in the training dynamics of large-scale multilingual SSL models. 

To summarize, our main contributions are as follows:
\begin{enumerate}
    \item We publicly release a new corpus that contains 7,413 hours of unlabeled speech across 4,057 languages, 25+ times wider coverage than current public datasets \cite{mlsuperb}.
    \item We introduce a new self-supervised task that improves model robustness by implicitly learning to clean reverberant audio.
    \item We publicly release XEUS, a SSL speech encoder trained on over 1 million hours of data across 4,057 languages. 
    \item We evaluate XEUS on numerous downstream tasks, and show that it outperforms SOTA SSL models such as MMS \cite{pratap2023scaling}, w2v-BERT 2.0 v2 \cite{barrault2023seamless}, and WavLM on tasks such as ASR, ST, and speech resynthesis.
    
\end{enumerate}
\begin{table*}[ht]
    \centering
    \caption{Comparison of multilingual SSL models by number of languages, training data size, and transparency. We define transparency in terms of the public availability of the data, model checkpoints (weights), training code and/or configurations, and the release of any other training artifacts (logs, intermediate checkpoints).}
   \resizebox{\textwidth}{!}{
    \begin{tabular}{l|rrcccc}
    \toprule
    Model  &  Langs. & Hours & \multicolumn{4}{c}{Transparency} \\
    & & & Data & Weights & Training Code & Other Artifacts \\
    \midrule
    XLS-R 128 \cite{babu2021xls} & 128 & 436K & \cmark & \cmark & \xmark & \xmark \\
    w2v-BERT 51 \cite{FLEURS} & 51 & 429K & \cmark & \xmark & \xmark & \xmark \\
    MR-HuBERT \cite{shi2024multiresolution} & 23 & 100K & \cmark & \cmark & \cmark & \xmark\\
    WavLabLM \cite{wavelablm} & 136 & 40K & \cmark & \cmark & \cmark & \xmark \\
    MMS \cite{pratap2023scaling} & 1,406 & 491K & \xmark & \cmark & \xmark & \xmark \\
    USM \cite{google-usm}& 300 & 12.5M & \xmark & \xmark & \xmark & \xmark \\
    w2v-BERT 2.0 v1 \cite{barrault2023seamlessm4t} & 143 & 1M & \xmark & \cmark & \xmark & \xmark\\
    w2v-BERT 2.0 v2 \cite{barrault2023seamless} & 143 & 4.5M & \xmark & \cmark & \xmark & \xmark\\
    \midrule
    XEUS (ours) & 4,057 & 1M & \cmark & \cmark & \cmark & \cmark \\
    \bottomrule
    \end{tabular}}
    \label{tab:models}
    \vspace{-0.4cm}
\end{table*}

\setlength{\textfloatsep}{6pt}
\begin{table}[htb]
    \centering
    \caption{Overview of datasets used for pre-training. The language column indicates the language used in monolingual datasets and the number of languages in multilingual datasets. \textbf{Bolded} dataset names indicate new corpora we will release.  }
    \vspace{-0.2cm}
    \resizebox{\columnwidth}{!}{
    \begin{tabular}{l|c|c|r}
    \toprule
    Dataset  &  Language(s) & Domain & Hours\\
    \midrule
    YODAS  \cite{yodas}  & 140 & Variety & 422K \\
    VoxPopuli \cite{voxpopuli} & 23 & Legal & 400K \\
    LibriLight \cite{kahnLibriLight} & English & Audiobook & 60K \\
    MLS \cite{pratap2020mls} & 8 & Audiobook & 44K \\
    People's Speech \cite{galvez2021people} & English & Variety & 30K \\
    WeNetSpeech \cite{wenetspeech} & Mandarin & Variety & 22K \\
    {Russian Open STT} \cite{ru-open-stt} & Russian & Variety & 20K \\
    {NPTEL2020} \cite{nptel} & English & Talk & 15K \\
    {Reazonspeech} \cite{reazonspeech} & Japanese & Television & 15K \\
    Common Voice 13 \cite{commonvoice} & 92 & Read & 13K \\
    GigaSpeech \cite{gigaspeech} & English & Variety & 10K \\
    VoxLingua  \cite{voxlingua} & 107 & Variety & 6800 \\
    \textbf{MMS-unlab v2} & 4023 & Religious & 6700 \\
    SPGI \cite{spgispeech} & English & Finance & 5000 \\
    Fisher \cite{fisher-callhome} & English & Conversation & 2000 \\
    Googlei18n \cite{wavelablm} & 34 & Variety & 1328\\
    {BABEL} \cite{babel} & 17 & Conversation & 1000 \\
    FLEURS \cite{FLEURS} & 102 & News & 1000 \\
    KSponSpeech \cite{ksponspeech} & Korean & Conversation & 970 \\
    LibriSpeech \cite{librispeech-corpus} & English & Audiobook & 960 \\
    MagicData \cite{magicdata} & Mandarin & Conversation & 755 \\
    mTEDx \cite{salesky2021mtedx} & 8 & Talk & 753 \\
    \textbf{Jesus Dramas} & 430 & Religious & 643 \\
    Althingi \cite{helgadottir19_althingi} & Icelandic & Legal & 542 \\
    TEDLIUM3 \cite{tedlium3} & English & Talk & 500 \\
    {VoxForge} \cite{voxforge} & 8 & Read & 235 \\
    AISHELL \cite{aishell-corpus} & Mandarin & Read & 200 \\
    SEAME  \cite{lyu10_seame} & Codeswitch & Conversation & 192 \\
    {DAMP-MVP} \cite{dampmvp} & English & Singing & 150 \\
    NorwegianParl. \cite{solberg-ortiz-2022-norwegian} & Norwegian & Legal & 140 \\
    {AIDATATANG} \cite{aidatatang} & Mandarin & Read & 140 \\
    AMI \cite{ami-corpus} & English & Meetings & 100 \\
    Nahuatl \cite{shi2021highland} & Nahuatl & Conversation & 82 \\
    WSJ \cite{wsj} & English & Read & 81 \\
    Mixtec \cite{shi2021mixtec} & Mixtec & Conversation & 70 \\
    \textbf{WikiTongues} & 700 & Conversation & 70 \\
    Siminchik \cite{cardenas2018siminchik} & Quechua & Radio & 50 \\ 
    Edinburgh Accent \cite{edinburgh} & English & Conversation & 40 \\
    {VCTK} \cite{vctk}  & English & Read & 25 \\
    AccentDB \cite{accentdb} & English & Read & 20 \\
    Totonac \cite{berrebbi22_totonac} & Totonac & Monologue & 17 \\
    \bottomrule
    \end{tabular}}
    \label{tab:data_part}
\end{table}

\vspace{-0.1cm}
\section{Motivation and Related Work}

\subsection{Speech Representation Learning}
SSL has seen tremendous success in speech processing by having neural networks learn rich feature representations from large-scale unlabeled data \cite{baevskiw2v, hsuHubert, ChenWavLm}, which can then be fine-tuned on various downstream tasks \cite{chenImproving, google-usm}. Multilingual SSL is a natural extension of this technique \cite{conneau2020unsupervised, babu2021xls} and facilitates cross-lingual transfer learning at scale. However, almost no studies leverage this capability to scale multilingual SSL models to truly massively multilingual settings, with the exception of Meta's MMS capable of covering $\sim$1,000+ languages \cite{pratap2023scaling}. However, MMS relies upon the older wav2vec 2.0 SSL objective, which has now been consistently outperformed by newer SSL objective \cite{superb, hsuHubert, chiu2022self}. In our work, we scale to 4 times the language coverage of MMS while further boosting performance with more powerful model architectures \cite{ebf} and training objectives \cite{ChenWavLm}. 

\subsection{Robust Speech Representations}
As most SSL speech encoders are trained solely on clean speech \cite{baevskiw2v, hsuHubert, chungW2vBert}, they perform noticeably worse on noisy audio \cite{changSSL}. A common approach to alleviate this issue is to perform continued pre-training of the SSL model in noisy conditions \cite{chang2023r, dehubert, distilldistort, robustd2v}. While computationally efficient, the performance of these methods is ultimately limited by the underlying SSL model. WavLM \cite{ChenWavLm} solves this issue by introducing an implicit denoising task during SSL pre-training, where the model must predict clean phonetic pseudo-labels when given an utterance corrupted with acoustic noise. WavLabLM \cite{wavelablm} extends this approach to the multilingual setting. Unlike XEUS, however, neither model considers the impact of reverberation, and both are trained on much smaller corpora (40K-86K hours vs.~1+ million hours).

\subsection{Open Foundation Models}
State-of-the-art speech foundation models vary significantly in their degree of openness (Table \ref{tab:models}). The best performing models like Whisper \cite{whisper}, Google USM \cite{google-usm}, w2v-BERT 2.0 v1 \cite{barrault2023seamlessm4t}, and w2v-BERT 2.0 v2 \cite{barrault2023seamless} are all trained on fully closed data. Whisper and w2v-BERT 2.0 v1/v2 only report pre-training data quantity and the languages covered. The USM report includes much more information about their data sources, but the model checkpoints remain unreleased.

Smaller scale multilingual speech models follow more open release practices. XLSR 53 \cite{conneau2020unsupervised} and XLS-R 128 \cite{babu2021xls} came with checkpoints and only use publicly accessible datasets but did not release training code. Similarly, MMS \cite{pratap2023scaling} released checkpoints but did not release their training code and crawled data. WavLabLM \cite{wavelablm} and MR-HuBERT \cite{shi2024multiresolution} released code and checkpoints but operated on a smaller scale.

Software infrastructure remains a critical barrier to democratizing speech SSL research. While there is plenty of infrastructure for large-scale training of text-based models \cite{workshop2022bloom, gpt-neox-library,liu2023llm360,groeneveld2024olmo}, no similar work has achieved speech pre-training at our scale. AR-HuBERT \cite{chen23l_interspeech} and the OWSM project \cite{asru23-owsm, peng2024owsm, peng2024owsmctc} sought to reproduce SOTA speech models in an open manner but use more than 80\% less data than our work. With XEUS, we release the entire pre-training framework. We only use publicly accessible datasets and release all of the additional pre-training data that we crawled. To facilitate research in the training dynamics of large-scale SSL models, we also release all model logs and are the first to also release all intermediate checkpoints. As we are the first to create an open SSL speech model at such data and model scale, we also release all of our heavily optimized training code. 

\begin{figure*}
    \centering
    \includegraphics[width=\textwidth]{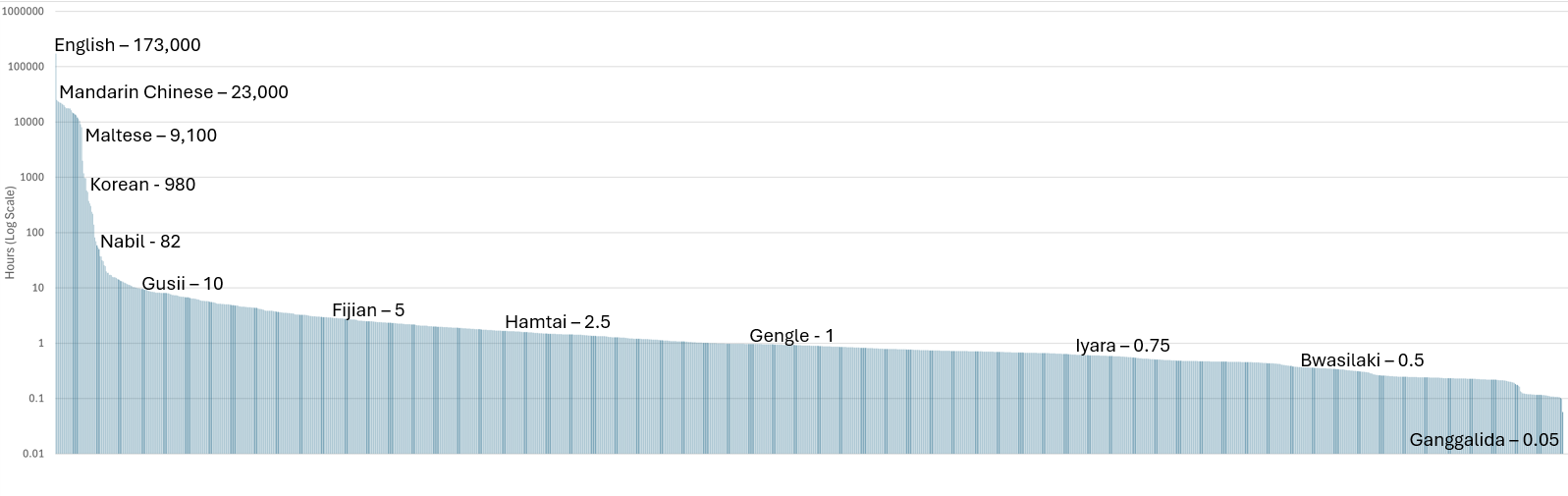}
    \caption{Distribution of XEUS pre-training data by language (log scale). We exclude data from YODAS \cite{yodas} due to the noisiness of the language labels.}
    \label{fig:lang_dist}
    \vspace{-0.4cm}
\end{figure*}

\vspace{-0.1cm}
\section{Data}
\vspace{-0.2cm}

\subsection{Existing Datasets}\label{sec:existing_data}
We begin by combining a large variety of pre-training data from 37 publicly accessible\footnote{We follow \citet{asru23-owsm}and include licensed data such as BABEL \cite{babel} as part of this definition, as exact copies can be obtained, unlike that of closed/unreleased data.} speech processing datasets across 150+ languages, which totals to 1.074 million hours of data. Details about these datasets can be found in Section \ref{sec:a_data} in the Appendix, while an overview is presented in Table \ref{tab:data_part}. Notable data used in this work that was unseen in prior SSL models includes accented speech \cite{accentdb, edinburgh}, code-switching \cite{lyu10_seame}, and singing voices \cite{dampmvp}. To prevent data corruption, we use only the official training splits of each dataset.  To increase language coverage beyond these 150 languages, in the next subsections we describe our collection of three additional data sources.

\vspace{-0.1cm}
\subsection{MMS-unlab v2}\label{sec:mms}
We first reproduce the MMS-unlab dataset \cite{pratap2023scaling}, which was not publicly released, and scale it to 200 more languages. Like the original, we crawl religious audiobooks from the {Global Recordings Network}.\footnote{https://globalrecordings.net/en/us} Our data, however, is processed in a different manner. Since we use it for SSL instead of language identification, we do not filter out languages with low amounts of data. We also perform VAD with an {energy-based detector}\footnote{https://github.com/wiseman/py-webrtcvad} instead of a neural model, the latter of which is more computationally expensive and likely less robust to unseen languages. This leads to a total of 6,700 hours of data across 4,023 ISO3 languages, which is wider in coverage than the original with 3,809 languages. We obtain explicit written permission for the use and redistribution of this data, as the Global Recordings Network website did not include clear licensing information. 

\vspace{-0.1cm}
\subsection{WikiTongues} \label{sec:nwikitongues}
\vspace{-0.1cm}
We also create new unlabeled speech corpora by crawling data from 2 data sources previously unseen in the speech processing literature. The first is {WikiTongues},\footnote{https://wikitongues.org} based on a grassroots project that collected recordings of many languages and dialects spoken around the world with the goal of language preservation. Each 2-20 minute recording is released under the CC-BY-NC license, and contains 1-2 speakers casually speaking a particular language/dialect. Importantly, many of these languages are low-resource, if not endangered or extinct. In total, the dataset we obtain contains around 700 languages and/or dialects. We obtained 821 recordings, yielding about 70 hours of speech data. We use the same VAD settings as in Section \ref{sec:mms} to segment each recording. 

\subsection{Jesus Dramas} \label{sec:jesusdramas}
The second corpus we collect is \textit{Jesus Dramas}. The source of this data is {Inspirational Films}\footnote{https://www.inspirationalfilms.com}, which released the ``Story of Jesus" audio drama in 430 languages under a non-commercial license. Each multi-speaker audio drama is 90 minutes long, totalling 645 hours. We use the same VAD settings as in Section \ref{sec:mms} to segment these dramas into utterance-level clips. 

\subsection{Final Pre-Training Corpus}
\vspace{-0.2cm}
The new datasets we collect from Sections \ref{sec:mms}, \ref{sec:nwikitongues} and \ref{sec:jesusdramas} total 7,413 hours of data across 4,057 ISO3 languages. After aggregating it with the data from Section \ref{sec:existing_data}, we obtain a total of 1.081 million hours of pre-training data. We filter out all utterances longer than 40 seconds due to memory constraints.  An overview of all of our pre-training corpora with their licensing information is presented in Table \ref{tab:data} in the Appendix. Figure \ref{fig:lang_dist} shows an overview of the language distribution in our data on a log-scale. Overall, our pre-training dataset spans 189 language families (Appendix Figure \ref{fig:families}). We find that the languages follow a long tail distribution, with the top 50 languages accounting for 99.5\% of the data. However, a very encouraging finding is that around 2,000 languages have 1 or more hours of data each. Data from YODAS (Appendix Section \ref{sec:a_data}) is excluded from the above analyses, as we found the language labels to be very noisy.

\section{Self-Supervised Pre-Training} \label{sec:ssl}
\begin{figure*}
    \centering
    \includegraphics[width=\textwidth]{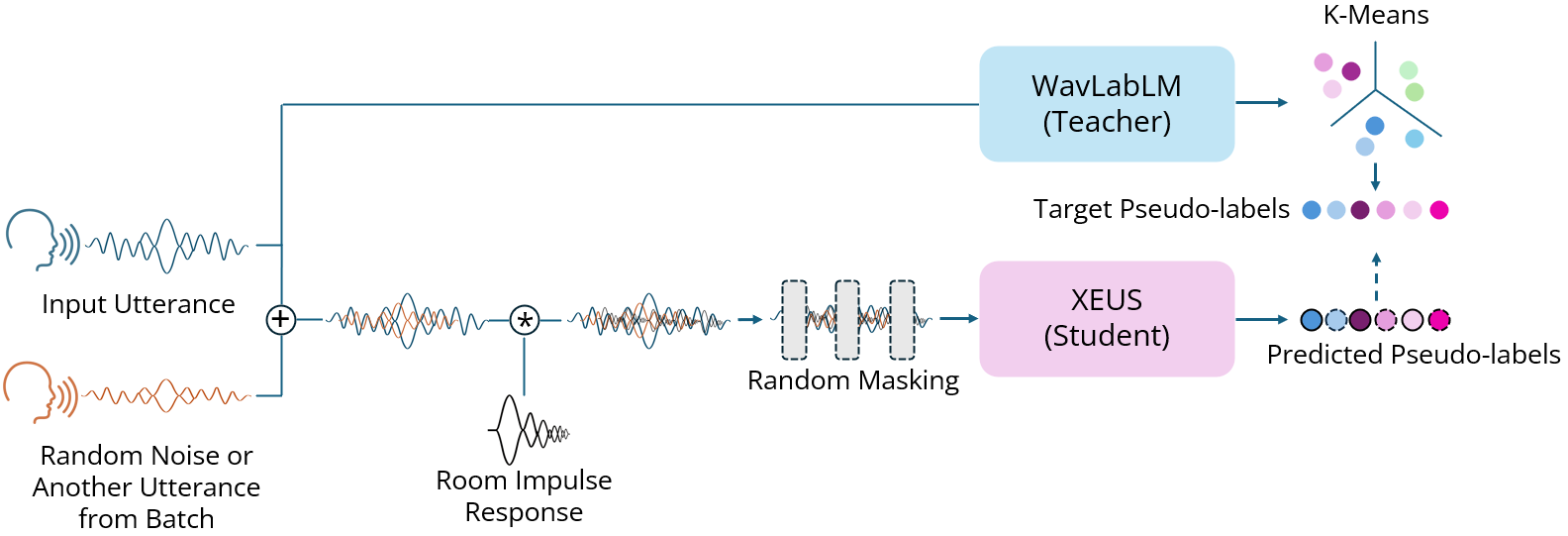}
    \caption{Overview of XEUS' pre-training. The teacher encoder generates phonetic pseudo-labels from clean speech, while the student must predict those pseudo-labels after masking, random noise and/or reverberation is applied to the input waveform.}
    \label{fig:diagram}
    \vspace{-0.4cm}
\end{figure*}

XEUS' training, sketched in Figure~\ref{fig:diagram}, combines ideas from HuBERT's masked prediction, WavLM's denoising objective, and a new dereverberation objective.  These components are described in the following sub-sections.

\subsection{Masked Prediction and Denoising} \label{sec:hubertwavlm}
To obtain the target phonetic pseudo-labels for HuBERT masked prediction, we first extract encoded representations from a pre-trained WavLabLM MS model~\cite{wavelablm}. The representations are then clustered using k-means, with $k=2,048$. The data used for the feature extraction and clustering is a subset of our training data. Specifically, we sample 6,000 hours from a combination of Common Voice, MLS, and Googlei18n, 6,000 hours from YODAS, and 6,000 hours from MMS-unlab v2, and combine it with the entirety of FLEURS and BABEL's training data. This leads to a total of 20K hours used for k-means.

We also integrate the acoustic denoising task proposed by WavLM into XEUS pre-training. During training, an input utterance has some probability $p$ to be augmented with either random noise from the Deep Noise Suppression Challenge \cite{dns_noise} or another utterance in the batch as interference. As the target labels are obtained solely from uncorrupted speech, the model learns to implicitly clean the input audio. 

\vspace{-0.3cm}
\subsection{Dereverberation} \label{sec:dereverb}
\vspace{-0.1cm}

\setlength{\textfloatsep}{2pt}
\begin{algorithm}[tb]
\caption{Simulation of Reverberant Speech}\label{alg:dereverb}
\begin{algorithmic}
\Require{ a batch of utterances $B$, 
a set of RIRs $D$, and 
reverberation probability $p_r$.}
\For{$u \in B$}
    \State \text{Sample $v$ from cont. dist. $\mathcal{U}(0,1)$}
    \If{$v < p_r$}
        \State \text{Sample a random RIR $u_n$ from $D$}
        \State{$dt = \min(\mathrm{argmax}(u_n))$}
        \State{$r = u \circledast u_n $}
        \State{Realign $r$ to $u$ using $dt$}
        \State{Rescale $r$ to have the same energy as $u$}
    \EndIf
\EndFor
\State \Return $B$
\end{algorithmic}
\end{algorithm}

We extend the concept of acoustic denoising introduced by WavLM by proposing another speech enhancement task for SSL pre-training: dereverberation. 
Similarly to the WavLM dynamic mixing augmentation, we simulate reverberant conditions in the input audio during training while the target pseudo-labels are again left untouched. The model must thus implicitly learn to remove the reverberation from the audio to predict the clean pseudo-labels. We note that it is possible for \textit{both} the noise and reverberation augmentation to be applied for a single utterance. For simplicity, the noise augmentation is always applied first.

Our technique (Algorithm \ref{alg:dereverb}) consists of the following steps. First, each utterance in a mini-batch has a probability $p_r$ for the reverberation augmentation to be applied. If an utterance $u$ is to be augmented, we then randomly sample a Room Impulse Response (RIR) $u_n$ to be used from the audio of \citet{rir_noises}. We first estimate $dt$, the sample shift imposed on $u$ after adding the reverberation, according to the highest peak in $u_n$. The reverberant utterance $r$ can then be simulated via convolution ($\circledast$) between $u$ and $u_n$. Finally, we realign $r$ with $u$ using $dt$ and normalize it to have the same energy as $u$. This final realignment step is crucial for the effectiveness of this technique; otherwise the audio would be shifted and misaligned with the target pseudo-labels. Quantitative analyses of our method can be found in Appendix Section \ref{sec:dereverb-ablation}.

\vspace{-0.3cm}
\subsection{Model Architecture} \label{sec:architecture}
\vspace{-0.1cm}
XEUS is based on the HuBERT architecture \cite{hsuHubert}, with several modifications. We replace the Transformer \cite{transformer} with an E-Branchformer \cite{branchformer,ebf}, as convolution-augmented models achieve superior SSL performance \cite{chungW2vBert}. We choose the E-Branchformer over the Conformer \cite{conformer} due to the former's relative ease of training and superior downstream performance \cite{ebf-vs-conformer}. We also replace the original HuBERT loss with cross entropy, which is faster and leads to better downstream performance \cite{fasthubert}. XEUS consists of a convolutional feature extractor and 19 E-Branchformer layers. Each of the latter has 8 attention heads, a hidden dimension of 1,024, feed-forward size of 4,096, and kernel size of 31. The model size is 577M parameters. Ablations on these modifications can be found in Appendix Section \ref{sec:ablation_arch}.

\vspace{-0.2cm}
\subsection{Pre-Training Settings} \label{sec:pretraining}
XEUS is pre-trained on 64 40GB NVIDIA A100 GPUs using the ESPnet toolkit \cite{espnet}. Each GPU has a maximum batch size of 100 seconds, leading to a total batch size of 106 minutes. We use a noise augmentation probability $p$ of 0.2, where there is an equal probability of the corruption being random noise or another utterance (Section \ref{sec:hubertwavlm}). We use a dereverberation augmentation probability $p_r$ of 0.3 (Section \ref{sec:dereverb}). We perform a two passes through the training set, totalling 670K steps. More details and a breakdown of the training costs can be found in Appendix Section \ref{a_pt}.

The scale of XEUS' pre-training is unseen outside of a few other works \cite{whisper, google-usm, barrault2023seamless}, with the amount of pre-training data being 5-25 times the size of those used in prior models trained with public toolkits \cite{asru23-owsm, chen23l_interspeech, hsuHubert}. To conduct training on such a scale, we made several optimizations to the open-source ESPnet toolkit, which was originally designed for standard academic-scale experiments. These optimizations will all be made publicly available. More details about these improvements are also reported in Appendix Section \ref{a_eng}.

\begin{table*}
\caption{Evaluation results on the 10 minute / 1 hour settings of the ML-SUPERB Benchmark in ASR CER ($\downarrow$) and LID ACC ($\uparrow$). \textbf{Bold} numbers indicate the best model for a task, while \underline{underlined} numbers indicate second best.} \label{tab:ml-superb-main}
    \resizebox {\textwidth} {!} {
\begin{tabular}{lcc|c|c|cc|c|ccc}
\toprule
\multirow{3}{*}{} & \multirow{3}{*}{} & \multirow{3}{*}{} & \multirow{3}{*}{} & \textsc{Mono. ASR} & \multicolumn{2}{c|}{\textsc{Multi. ASR}} & \multicolumn{1}{c|}{\textsc{LID}} & \multicolumn{3}{c}{\textsc{Multi. ASR + LID}} \\
\cmidrule(lr){5-5} \cmidrule(lr){6-7} \cmidrule(lr){8-8} \cmidrule(lr){9-11}
& &    &      &     &        Normal & Few-shot & Normal & \multicolumn{2}{c}{Normal} & \multicolumn{1}{c}{Few-shot} \\
Model & Params. & Hours & SUPERB$_{s}$ & CER & CER & CER & ACC & ACC & CER & \multicolumn{1}{c}{CER}  \\
\midrule
XLS-R 128  & 316M & 436K & 707 / 851 & 39.7 / 30.6 & 29.3 / 22.0 & 40.9 / 39.3 & 66.9 / \textbf{87.9} & 55.6 / 85.6 & 28.4 / 22.9 & 42.1 / 42.4  \\
XLS-R 128  & 1B & 436K & 745 / 838 & 40.5 / 30.9 & 30.4 / 26.3 &  39.1 / 38.5 & 70.9 / 85.8  &  66.4 / 87.1 & 28.6 / 25.2 & 39.2 / 39.5 \\
MMS  & 316M & 491K & 795 / 845 & 33.8 / 30.5 & 28.7 / 24.0 & 36.5 / 36.5 & 62.3 / 84.3 & 71.9 / 74.3 & 31.5 / 30.0 & \underline{30.9} / \underline{29.2} \\
MMS  & 1B  & 491K & \underline{953} / \underline{948} & \underline{33.3} / \underline{25.7} & \underline{21.3} / \textbf{18.1} & \textbf{30.2} / \textbf{30.8} & \textbf{84.8} / 86.1 & 73.3 / 74.8 & 26.0 / 25.5  & \textbf{25.4} / \textbf{24.8} \\
w2v-BERT 2.0 v2  & 580M & 4.5M & 826 / 916 & 41.0 / 29.2 & 24.6 / 20.3 & 34.3 / 35.3 & 70.0 / 86.8 & \underline{83.2} / \underline{90.6} &  \underline{24.2} / \underline{20.3} & 34.3 / 34.0\\
\midrule
XEUS (ours) & 577M & 1M &  \textbf{956} / \textbf{956} & \textbf{30.3 }/ \textbf{25.1} & \textbf{21.1} / \underline{20.1} &  \underline{33.4} / \underline{34.1} & \underline{81.5} / \underline{87.3} & \textbf{86.4} / \textbf{91.3} & \textbf{22.9} / \textbf{19.6} & 32.7 / 32.8\\
\bottomrule
\end{tabular}}
\vspace{-0.4cm}
\end{table*}

\begin{table}[ht]
    \centering
    \caption{FLEURS ASR+LID \& JesusFilm ST results. }
    \vspace{-0.2cm}
    \resizebox {\columnwidth} {!} {
    \begin{tabular}{l|cc|c}
    \toprule
     Model   & \multicolumn{2}{c|}{FLEURS} & JesusFilm \\
    & CER $\downarrow$ & ACC $\uparrow$ & chrF $\uparrow$ \\
     \midrule
     XLS-R 1B & 9.6 & 92.5 & 13.4 \\
     MMS 1B  & 9.2 & 94.0 & 15.5 \\
     w2v-BERT 2.0  v2  & \textbf{8.7} & \textbf{94.3} & 15.1\\
     \midrule
     XEUS (ours)  &  8.9 & 93.0 & \textbf{22.1} \\
     \bottomrule
    \end{tabular}}
    \label{tab:fleurs}
    \vspace{-0.1cm}
\end{table}

\vspace{-0.2cm}
\section{Downstream Evaluation}
\vspace{-0.1cm}
We examine the capabilities of XEUS in various downstream applications. Section \ref{masr} evaluates the multilingual capabilities of XEUS in different settings. Section \ref{SUPERB} evaluates the universality of XEUS' representations for a broad range of speech information, such as emotion and speaker content. Finally, Section \ref{generation} tests the acoustic representations of XEUS via speech resynthesis. We provide an overview of each downstream setup in its respective subsection. Detailed experimental settings can be found in Appendix Section \ref{a_exp}.

\vspace{-0.1cm}
\subsection{Multilingual Speech Processing} \label{masr} 
We primarily compare XEUS with 3 SOTA multilingual SSL models: XLS-R 128 \cite{babu2021xls}, MMS \cite{pratap2023scaling}, and w2v-BERT 2.0 v2 \cite{barrault2023seamless} (Table \ref{tab:models}). We use the ML-SUPERB benchmark \cite{mlsuperb} as our main evaluation, as it tests each models across a diverse range of tasks and languages. We complement these experiments with additional analyses on FLEURS ASR+LID and low-resource ST.

\vspace{-0.2cm}
\subsubsection{ML-SUPERB} \label{ml-superb}
ML-SUPERB benchmarks self-supervised speech representations on a variety of multilingual tasks across 143 languages. ASR performance is evaluated in terms of character error rate (CER $\downarrow$), while accuracy (ACC $\uparrow$) is used to evaluate LID. The benchmark is split across two data settings for each task: 10 minutes (min.) and 1 hour of data for each language. Each data setting contains 4 tasks: monolingual ASR in 9 languages, multilingual ASR, LID, and joint multilingual ASR+LID. In the multilingual tasks, 5 languages are reserved as a few-shot task, while the other 138 languages have the standard 10 min. / 1 hour of fine-tuning data. An overall SUPERB$_{s}(\uparrow)$ score for each model in each data setting is calculated following the benchmark rules \cite{mlsuperb}. Further details can be found in Appendix Section \ref{sec:a_mlsuperb}.

Table \ref{tab:ml-superb-main} shows that XEUS is the overall best performing model, with the highest SUPERB$_s$ score of 956 on both the 10 min. / 1 hour settings. XEUS achieves SOTA results on monolingual ASR with the best scores of 25.1 and 33.3 CER on the 1 hour and 10 min. tracks respectively. On multilingual ASR, XEUS is only outperformed by MMS 1B. For ASR+LID, XEUS achieves the best performance in the normal setting for both data tracks. While XEUS is worse than MMS in few-shot CER, it still achieves reasonable results and outperforms the other SSL models. Overall, XEUS outperforms the parameter-equivalent w2v-BERT 2.0 v2 across all task categories. This is accomplished using only accessible training data, which is 22\% the size than that of w2v-BERT 2.0 v2.

\setlength{\textfloatsep}{2pt}
\begin{table*}[ht]
\caption{Evaluation on the SUPERB Benchmark. (\emojifirst), (\emojisecond), and (\emojithird) indicate first, second and third best results respectively on the online leaderboard: \url{https://superbbenchmark.org/leaderboard}.} \label{tab:superb}
 \resizebox {\linewidth} {!} {
 \begin{tabular}{lccccccccccc}
 \toprule
 \multirow{3}*[-2pt]{Method} & \multicolumn{2}{c}{Recognition} & \multicolumn{2}{c}{Detection}  & \multicolumn{3}{c}{Semantics} & \multicolumn{3}{c}{Speaker} & Paralinguistics \\
 \cmidrule(r){2-3}\cmidrule(r){4-5}\cmidrule(r){6-8}\cmidrule(r){9-11}\cmidrule(r){12-12}
 & PR$\downarrow$ & ASR$\downarrow$ & KS $\uparrow$ & QbE $\uparrow$  & IC $\uparrow$ & SF (F1) $\uparrow$ & SF (CER) $\downarrow$ & SID $\uparrow$ & ASV $\downarrow$  & SD $\downarrow$ & ER $\uparrow$ \\
 \midrule
 WavLM Large   & \textbf{3.06} \emojifirst & 3.44 \emojithird & 97.86 \emojisecond & \textbf{8.86}   & \textbf{99.31} \emojifirst & \textbf{92.21} \emojifirst & \textbf{18.36} \emojifirst & \textbf{95.49} \emojifirst & \textbf{3.77} \emojifirst & 3.24 \emojisecond & 70.62 \emojisecond \\
\midrule
XEUS (ours) & 3.21 \emojithird & \textbf{3.34} \emojifirst  & \textbf{98.32} \emojifirst & 7.49  & 98.70 \emojinone & 90.05 \emojinone & 21.49 \emojithird & 91.70 \emojisecond & 4.16 \emojithird & \textbf{3.11} \emojifirst & \textbf{71.08} \emojifirst \\
\bottomrule
 \end{tabular}
 }
 \vspace{-0.4cm}
 \end{table*}

\setlength{\textfloatsep}{2pt}
\begin{table}[tb]

    \centering
    \caption{Speech resynthesis results on VCTK.}
    \label{tab:resynth}
    \resizebox {\columnwidth} {!} {
    \begin{tabular}{lcccc}
    \toprule
    Model & MOS $(\uparrow)$ & WER $(\downarrow)$ & F0 $(\downarrow)$ & MCD  $(\downarrow)$\\
    \midrule
    WavLM Large & 3.20 & 27.8 & 0.26 & 4.55\\
    w2v-BERT 2.0  v2  & 3.21 & 15.5 & 0.27 & 3.92 \\
    \midrule
    XEUS (ours)  & \textbf{3.23} & \textbf{10.0} & \textbf{0.25} & \textbf{3.80}\\ 
    \bottomrule
    \end{tabular}}
\end{table}

\vspace{-0.15cm}
\subsubsection{FLEURS}\label{sec:fleurs}
FLEURS is a 102-language multilingual ASR benchmark, where each language has around 6-10 hours of training data. In this setting, we use heavier downstream probes that reflect SOTA ASR architectures. We adopt the same setup as \cite{ebf-vs-conformer, chenImproving}, which remains the SOTA on FLEURS when not using additional labeled data. The downstream model consists of an E-Branchformer encoder paired with a Transformer decoder, totalling 100M parameters. Exact settings are shown in Appendix Section \ref{sec:a_fleurs}.

The results of the FLEURS experiments are shown in the middle section of Table \ref{tab:fleurs}. We find that XEUS remains competitive with the SOTA w2v-BERT 2.0 v2 trained on much more data (8.9 vs 8.7 CER), and significantly outperforms both XLS-R and MMS 1B (9.6 and 9.2 CER respectively). 

\vspace{-0.15cm}
\subsubsection{Low-Resource Language Coverage} \label{st}
While FLEURS and ML-SUPERB provide comprehensive multilingual benchmarks, their language coverage is far smaller than that of XEUS (102-143 vs 4,057). To understand if XEUS' wide language coverage was effective for languages with small (< 10 hours) amounts of pre-training data, we crawled additional labeled data for evaluation. We randomly chose 3 languages from the {Jesus Film Project}\footnote{https://www.jesusfilm.org} that were not covered in ML-SUPERB and/or FLEURS: Hijazi Arabic, Lumun, and Rajbanshi.  This yields around 1.5 hours of speech for each language. Of all existing SSL models, only XEUS covers the former two, while Rajbanshi is covered by both XEUS and MMS. We then train X $\rightarrow$ English Speech Translation (ST) models for each language (as ASR transcripts were not available) and evaluate using character-level F-score (chrF). More details about the evaluation data and ST settings are shown in Appendix Section \ref{sec:a_st}.

The average chrF scores across all languages are shown in the right of Table \ref{tab:fleurs}. XEUS significantly outperforms all other models, likely due to its wider language coverage. The next best model is MMS 1B, which obtains an average chrF of 15.5, while XEUS scores 22.1, a relative improvement of 35\%.

\vspace{-0.2cm}
\subsection{Task Universality} \label{SUPERB}
To test how well XEUS encodes other forms of speech content, we benchmark its capabilities on the English-only SUPERB \cite{superb}. SUPERB tests self-supervised speech models across 5 broad task categories: Recognition (ASR and Phoneme Recognition (PR)), Detection (Keyword Spotting (KS) and Query By Example (QbE)), Semantics (Intent Classification (IC) and Slot Filling (SF)), Speaker (Speaker Identification (SID), Automatic Speaker Verification (ASV), and Speaker Diarization (SD)), and Paralinguistics (Emotion Recognition (ER)). Metrics for each task are: phoneme error rate (PR), WER (ASR), maximum term weighted value (QbE), F1 and concept error rate (SF), equal error rate (ASV), diarization error rate (SD), and accuracy (KS, IC, SID, and ER). Exact experimental settings are shown in Appendix Section \ref{sec:a_superb}. We compare XEUS to WavLM  \cite{ChenWavLm}, the SOTA model on the SUPERB leaderboard for almost all tasks. Our results in Table \ref{tab:superb} show that XEUS consistently reaches if not surpasses SOTA scores across a variety of tasks, obtaining the highest score in 4 English-only tasks (KS, SD, ER, ASR), despite its curse of multilinguality \cite{conneau-etal-2020-unsupervised}.

\vspace{-0.1cm}
\subsection{Acoustic Representation Evaluation} \label{generation}
\vspace{-0.1cm}
We evaluate XEUS on its acoustic representation quality through the task of speech resynthesis. Here, we compare primarily against w2v-BERT 2.0 v2 as the SOTA multilingual model and WavLM Large as the SOTA English-only model. We train unit-to-speech HiFiGAN vocoders \cite{hifigan, resynthesis} on the accented-English VCTK \cite{vctk} dataset with a discrete codebook vocabulary size of 100. We evaluate the quality of the resynthesized speech in terms of Mel-Cepstral Distortion (MCD), log-F0 Root Mean Square Error (F0), predicted Mean Opinion Score \cite{mosnet} (MOSNet), and Word Error Rate (WER). We obtain WER by transcribing the synthesized speech with a pre-trained Whisper medium \cite{whisper}. For each SSL model, we experiment with features extracted at 50\% and 75\% of the model depth (ie. layer 18 out of 24 in the latter case), and report the best performing configuration in Table \ref{tab:resynth}. More details about this search can be found in Appendix Section \ref{a_exp}. Our results show that resynthesized speech from XEUS is higher quality than that from both WavLM and w2v-BERT 2.0 v2 across all metrics, whether it be perceptual or semantic, showcasing its strong performance in generative tasks along with its SOTA recognition capabilities.

\vspace{-0.2cm}
\section{Conclusion}
\vspace{-0.2cm}
This work presents XEUS, an SSL speech encoder trained on over 1 million hours of data across 4,057 languages. As a community contribution, we release a new dataset with 7,413 hours of unlabeled speech data across those 4,057 languages. We also introduce a novel joint dereverberation task for SSL pre-training, which we use to increase the robustness of XEUS. We show that XEUS can achieve comparable performance if not outperform other SOTA SSL models on various benchmarks, while having much stronger performance on long-tail languages. To make XEUS reproducible, we will release all training code and configurations, along with model weights. In the future, we hope to extend the downstream use of XEUS to a larger scale.


\section{Limitations: }
While the overall pre-training corpus of XEUS contains over 4,000 languages, many of these languages have less than 1 hour of speech data. While we are able to show that the presence of this small amount of data is still beneficial for these languages in downstream tasks (Section \ref{st}), the performance is still likely much worse than the performance on higher-resourced languages. Furthermore, due to the efforts required to collect and manually clean evaluation data, we only test on a subset of these truly low-resource languages in Section \ref{st}. While XEUS is one step towards speech recognition or translation systems for these tail languages, much work is still required before these tools can be deployed to end users.

Due to the large number of tasks and domains that our evaluation covers, we mostly focus on relatively lightweight benchmarks such as the SUPERB suite and perform limited hyperparameter tuning. While this allows for fair comparisons between different models, the evaluation does not use the large-scale fine-tuning common in SOTA settings for downstream tasks. 

\section{Broader Impact and Ethics: }
\noindent \textbf{Broader Impact: } 
In this work, we introduce XEUS, a new large-scale multilingual SSL speech encoder. Unlike previous foundation models that focus on a single aspect, XEUS obtains SOTA performance across a diverse range of \textit{both} tasks \textit{and} languages, further pushing towards the goal of truly universal models. By releasing both our data and model checkpoints, our goal is to provide foundations for more multilingual research, particularly in domains such as robust ASR and speech enhancement where evaluation is typically done solely on English.

Another major goal of our work is to democratize the development of speech foundation models. We believe that training infrastructure remains a significant barrier to entry for SSL research. This has two main aspects: \textit{software infrastructure} and \textit{training configurations}. Current speech toolkits such as ESPnet \cite{espnet} and SpeechBrain \cite{ravanelli2021speechbrain} focus on academic scale experiments, while general frameworks such as HuggingFace and Fairseq \cite{wang2020fairseq} are more limited in their implementation of different tasks and SOTA methods. By integrating our changes into ESPnet, our optimizations can allow users to scale speech models for other tasks such speaker representation learning. In the latter aspect, we note that the availability of training recipes and configurations pose the other major barrier to entry. Due to the computational cost of training, the development of foundation models poses a risk that is too high for most academic labs, as a single failed training run can be disastrous for the lab's budget. However, this can be mitigated by publishing training recipes and hyperparameters known to work well. The benefits of this is most visible in the OWSM project \cite{asru23-owsm, peng2024owsm, peng2024owsmctc}, where each successive work reported lower and lower computational expenses.

\noindent \textbf{Ethics: } We recognize the delicate nature of speech data, particularly when it involves the languages of indigenous and marginalized communities. Many authors of this work have long-term collaborations with indigenous communities and researchers. We are careful to crawl and release data only from sources that contain permissive licenses of the source data to avoid potential cases of misuse and violations of language ownership. For data sources that do not clearly indicate their usage/distribution terms, we obtained explicit permission from the corresponding stakeholders (such as in the case of the Global Recordings Network in Section \ref{sec:mms}). To follow the data's access conditions, we release all of our data under non-commercial licenses.

We partially anonymize our crawled datasets by removing speaker names from the meta-data. However, we do not alter the content of the speech itself. As such, we urge users of our released data to respect the privacy of the speakers and not attempt to identify them. It is also possible that portions of the speech content may be offensive in particular settings. With the diversity of over 4000 languages, it is likely that there are statements or views that are normative in one culture but offensive in another.

While encoder-only speech models like XEUS have limited uses without any task-specific fine-tuning, the downstream models that are created after such processes are prone to the biases and misuse cases that all machine learning models are vulnerable to. For example, XEUS' capabilities in speech generation can be used for misinformation via audio deepfakes, which is an unintended use case of this model. 

\section*{Acknowledgement}
This work used the Bridges2 at PSC and Delta NCSA computing systems through allocation CIS210014 from the Advanced Cyberinfrastructure Coordination Ecosystem: Services \& Support (ACCESS) program, supported by National Science Foundation grants 2138259,2138286, 2138307, 2137603, and 2138296.

\bibliography{custom}

\begin{thebibliography}{96}
\providecommand{\natexlab}[1]{#1}

\bibitem[{aid()}]{aidatatang}

\newblock {aidatatang\_200zh, a free Chinese Mandarin speech corpus by Beijing DataTang Technology Co., Ltd}.

\bibitem[{Ahamad et~al.(2020)Ahamad, Anand, and Bhargava}]{accentdb}
Afroz Ahamad, Ankit Anand, and Pranesh Bhargava. 2020.
\newblock Accent{DB}: A database of non-native {E}nglish accents to assist neural speech recognition.
\newblock In \emph{LREC 2020}.

\bibitem[{{AI4Bharat}(2020)}]{nptel}
{AI4Bharat}. 2020.
\newblock \href {https://github.com/AI4Bharat/NPTEL2020-Indian-English-Speech-Dataset} {{NPTEL2020: Indian English Speech Dataset}}.

\bibitem[{Andonian et~al.(2023)Andonian, Anthony, Biderman, Black, Gali, Gao, Hallahan, Levy-Kramer, Leahy, Nestler, Parker, Pieler, Phang, Purohit, Schoelkopf, Stander, Songz, Tigges, Thérien, Wang, and Weinbach}]{gpt-neox-library}
Alex Andonian, Quentin Anthony, Stella Biderman, Sid Black, Preetham Gali, Leo Gao, Eric Hallahan, Josh Levy-Kramer, Connor Leahy, Lucas Nestler, Kip Parker, Michael Pieler, Jason Phang, Shivanshu Purohit, Hailey Schoelkopf, Dashiell Stander, Tri Songz, Curt Tigges, Benjamin Thérien, Phil Wang, and Samuel Weinbach. 2023.
\newblock \href {https://doi.org/10.5281/zenodo.5879544} {{GPT-NeoX: Large Scale Autoregressive Language Modeling in PyTorch}}.

\bibitem[{Ardila et~al.(2020)Ardila, Branson, Davis, Kohler, Meyer, Henretty, Morais, Saunders, Tyers, and Weber}]{commonvoice}
Rosana Ardila, Megan Branson, Kelly Davis, Michael Kohler, Josh Meyer, Michael Henretty, Reuben Morais, Lindsay Saunders, Francis Tyers, and Gregor Weber. 2020.
\newblock Common voice: A massively-multilingual speech corpus.
\newblock In \emph{LREC 2020}, pages 4218--4222.

\bibitem[{Austin and Sallabank(2011)}]{Austin2011-AUSTCH-2}
Peter~K. Austin and Julia Sallabank, editors. 2011.
\newblock \emph{The Cambridge Handbook of Endangered Languages}.
\newblock Cambridge University Press.

\bibitem[{Babu et~al.(2022)Babu, Wang, Tjandra, Lakhotia, Xu, Goyal, Singh, {von Platen}, Saraf, Pino, Baevski, Conneau, and Auli}]{babu2021xls}
Arun Babu, Changhan Wang, Andros Tjandra, Kushal Lakhotia, Qiantong Xu, Naman Goyal, Kritika Singh, Patrick {von Platen}, Yatharth Saraf, Juan Pino, Alexei Baevski, Alexis Conneau, and Michael Auli. 2022.
\newblock \href {https://doi.org/10.21437/Interspeech.2022-143} {{XLS-R: Self-supervised Cross-lingual Speech Representation Learning at Scale}}.
\newblock In \emph{Interspeech 2022}, pages 2278--2282.

\bibitem[{Baevski et~al.(2022)Baevski, Hsu, Xu, Babu, Gu, and Auli}]{baevski2022data2vec}
Alexei Baevski, Wei-Ning Hsu, Qiantong Xu, Arun Babu, Jiatao Gu, and Michael Auli. 2022.
\newblock Data2vec: A general framework for self-supervised learning in speech, vision and language.
\newblock In \emph{ICML 2022}, pages 1298--1312.

\bibitem[{Baevski et~al.(2020)Baevski, Zhou, Mohamed, and Auli}]{baevskiw2v}
Alexei Baevski, Yuhao Zhou, Abdelrahman Mohamed, and Michael Auli. 2020.
\newblock {wav2vec} 2.0: A framework for self-supervised learning of speech representations.
\newblock In \emph{NeurIPS 2020}, volume~33.

\bibitem[{Bang et~al.(2020)}]{ksponspeech}
Jeong-Uk Bang et~al. 2020.
\newblock {K}spon{S}peech: Korean spontaneous speech corpus for automatic speech recognition.
\newblock \emph{Applied Sciences}.

\bibitem[{Barrault et~al.(2023{\natexlab{a}})Barrault, Chung, Meglioli, Dale, Dong, Duquenne, Elsahar, Gong, Heffernan, Hoffman et~al.}]{barrault2023seamlessm4t}
Lo{\"\i}c Barrault, Yu-An Chung, Mariano~Cora Meglioli, David Dale, Ning Dong, Paul-Ambroise Duquenne, Hady Elsahar, Hongyu Gong, Kevin Heffernan, John Hoffman, et~al. 2023{\natexlab{a}}.
\newblock {SeamlessM4T}-massively multilingual \& multimodal machine translation.
\newblock \emph{arxiv:2308.11596}.

\bibitem[{Barrault et~al.(2023{\natexlab{b}})Barrault, Chung, Meglioli, Dale, Dong, Duppenthaler, Duquenne, Ellis, Elsahar, Haaheim et~al.}]{barrault2023seamless}
Lo{\"\i}c Barrault, Yu-An Chung, Mariano~Coria Meglioli, David Dale, Ning Dong, Mark Duppenthaler, Paul-Ambroise Duquenne, Brian Ellis, Hady Elsahar, Justin Haaheim, et~al. 2023{\natexlab{b}}.
\newblock Seamless: Multilingual expressive and streaming speech translation.
\newblock \emph{arxiv:2312.05187}.

\bibitem[{Berrebbi et~al.(2022)Berrebbi, Shi, Yan, López-Francisco, Amith, and Watanabe}]{berrebbi22_totonac}
Dan Berrebbi, Jiatong Shi, Brian Yan, Osbel López-Francisco, Jonathan Amith, and Shinji Watanabe. 2022.
\newblock \href {https://doi.org/10.21437/Interspeech.2022-10796} {Combining spectral and self-supervised features for low resource speech recognition and translation}.
\newblock In \emph{Interspeech 2022}.

\bibitem[{Black(2019)}]{wilderness}
Alan~W Black. 2019.
\newblock \href {https://doi.org/10.1109/ICASSP.2019.8683536} {{CMU Wilderness Multilingual Speech Dataset}}.
\newblock In \emph{ICASSP 2019}.

\bibitem[{Bu et~al.(2017)}]{aishell-corpus}
Hui Bu et~al. 2017.
\newblock {AISHELL-1: An open-source Mandarin speech corpus and a speech recognition baseline}.
\newblock In \emph{O-COCOSDA}.

\bibitem[{Cardenas et~al.(2018)Cardenas, Zevallos, Baquerizo, and Camacho}]{cardenas2018siminchik}
Ronald Cardenas, Rodolfo Zevallos, Reynaldo Baquerizo, and Luis Camacho. 2018.
\newblock Siminchik: A speech corpus for preservation of southern {Quechua}.
\newblock \emph{ISI-NLP}.

\bibitem[{Carletta(2007)}]{ami-corpus}
Jean Carletta. 2007.
\newblock Unleashing the killer corpus: experiences in creating the multi-everything {AMI} meeting corpus.
\newblock Springer.

\bibitem[{Chang and Glass(2023)}]{chang2023r}
Heng-Jui Chang and James Glass. 2023.
\newblock R-spin: Efficient speaker and noise-invariant representation learning with acoustic pieces.
\newblock \emph{arXiv preprint arXiv:2311.09117}.

\bibitem[{Chang et~al.(2021)Chang, Maekaku, Guo, Shi, Lu, Subramanian, Wang, Yang, Tsao, Lee, and Watanabe}]{changSSL}
Xuankai Chang, Takashi Maekaku, Pengcheng Guo, Jing Shi, Yen-Ju Lu, Aswin~Shanmugam Subramanian, Tianzi Wang, Shu-wen Yang, Yu~Tsao, Hung-yi Lee, and Shinji Watanabe. 2021.
\newblock \href {https://doi.org/10.1109/ASRU51503.2021.9688137} {An exploration of self-supervised pretrained representations for end-to-end speech recognition}.
\newblock In \emph{ASRU 2021}, pages 228--235.

\bibitem[{Chen et~al.(2021)}]{gigaspeech}
Guoguo Chen et~al. 2021.
\newblock {GigaSpeech}: An evolving, multi-domain {ASR} corpus with 10,000 hours of transcribed audio.
\newblock In \emph{Interspeech 2021}.

\bibitem[{Chen et~al.(2022)Chen, Wang, Chen, Wu, Liu, Chen, Li, Kanda, Yoshioka, Xiao, Wu, Zhou, Ren, Qian, Qian, Wu, Zeng, Yu, and Wei}]{ChenWavLm}
Sanyuan Chen, Chengyi Wang, Zhengyang Chen, Yu~Wu, Shujie Liu, Zhuo Chen, Jinyu Li, Naoyuki Kanda, Takuya Yoshioka, Xiong Xiao, Jian Wu, Long Zhou, Shuo Ren, Yanmin Qian, Yao Qian, Jian Wu, Michael Zeng, Xiangzhan Yu, and Furu Wei. 2022.
\newblock \href {https://doi.org/10.1109/JSTSP.2022.3188113} {{WavLM}: Large-scale self-supervised pre-training for full stack speech processing}.
\newblock \emph{IEEE JSTSP}.

\bibitem[{Chen et~al.(2023{\natexlab{a}})Chen, Chang, Peng, Ni, Maiti, and Watanabe}]{chen23l_interspeech}
William Chen, Xuankai Chang, Yifan Peng, Zhaoheng Ni, Soumi Maiti, and Shinji Watanabe. 2023{\natexlab{a}}.
\newblock {Reducing Barriers to Self-Supervised Learning: {HuBERT} Pre-training with Academic Compute}.
\newblock In \emph{Interspeech 2023}.

\bibitem[{Chen et~al.(2023{\natexlab{b}})Chen, Shi, Yan, Berrebbi, Zhang, Peng, Chang, Maiti, and Watanabe}]{wavelablm}
William Chen, Jiatong Shi, Brian Yan, Dan Berrebbi, Wangyou Zhang, Yifan Peng, Xuankai Chang, Soumi Maiti, and Shinji Watanabe. 2023{\natexlab{b}}.
\newblock Joint prediction and denoising for large-scale multilingual self-supervised learning.
\newblock In \emph{ASRU 2023}.

\bibitem[{Chen et~al.(2023{\natexlab{c}})Chen, Yan, Shi, Peng, Maiti, and Watanabe}]{chenImproving}
William Chen, Brian Yan, Jiatong Shi, Yifan Peng, Soumi Maiti, and Shinji Watanabe. 2023{\natexlab{c}}.
\newblock Improving massively multilingual {ASR} with auxiliary {CTC} objectives.
\newblock In \emph{ICASSP 2023}.

\bibitem[{Chiu et~al.(2022)Chiu, Qin, Zhang, Yu, and Wu}]{chiu2022self}
Chung-Cheng Chiu, James Qin, Yu~Zhang, Jiahui Yu, and Yonghui Wu. 2022.
\newblock Self-supervised learning with random-projection quantizer for speech recognition.
\newblock In \emph{International Conference on Machine Learning}. PMLR.

\bibitem[{Chung et~al.(2021)Chung, Zhang, Han, Chiu, Qin, Pang, and Wu}]{chungW2vBert}
Yu-An Chung, Yu~Zhang, Wei Han, Chung-Cheng Chiu, James Qin, Ruoming Pang, and Yonghui Wu. 2021.
\newblock \href {https://doi.org/10.1109/ASRU51503.2021.9688253} {w2v-{BERT}: Combining contrastive learning and masked language modeling for self-supervised speech pre-training}.
\newblock In \emph{ASRU 2021}.

\bibitem[{Conneau et~al.(2021)Conneau, Baevski, Collobert, Mohamed, and Auli}]{conneau2020unsupervised}
Alexis Conneau, Alexei Baevski, Ronan Collobert, Abdelrahman Mohamed, and Michael Auli. 2021.
\newblock \href {https://doi.org/10.21437/Interspeech.2021-329} {{Unsupervised Cross-Lingual Representation Learning for Speech Recognition}}.
\newblock In \emph{Interspeech 2021}, pages 2426--2430.

\bibitem[{Conneau et~al.(2020)Conneau, Khandelwal, Goyal, Chaudhary, Wenzek, Guzm{\'a}n, Grave, Ott, Zettlemoyer, and Stoyanov}]{conneau-etal-2020-unsupervised}
Alexis Conneau, Kartikay Khandelwal, Naman Goyal, Vishrav Chaudhary, Guillaume Wenzek, Francisco Guzm{\'a}n, Edouard Grave, Myle Ott, Luke Zettlemoyer, and Veselin Stoyanov. 2020.
\newblock \href {https://doi.org/10.18653/v1/2020.acl-main.747} {Unsupervised cross-lingual representation learning at scale}.
\newblock In \emph{ACL 2020}, pages 8440--8451, Online. Association for Computational Linguistics.

\bibitem[{Conneau et~al.(2022)}]{FLEURS}
Alexis Conneau et~al. 2022.
\newblock {FLEURS}: Few-shot learning evaluation of universal representations of speech.
\newblock In \emph{SLT 2022}.

\bibitem[{Dao et~al.(2022)Dao, Fu, Ermon, Rudra, and R{\'{e}}}]{flashattn}
Tri Dao, Daniel~Y. Fu, Stefano Ermon, Atri Rudra, and Christopher R{\'{e}}. 2022.
\newblock Flash{A}ttention: Fast and memory-efficient exact attention with {IO}-awareness.
\newblock In \emph{NeurIPS 2022}.

\bibitem[{{Ethnologue}(2017)}]{ethnologue}
{Ethnologue}. 2017.
\newblock \href {https://web.archive.org/web/20170227130129/https://www.ethnologue.com/enterprise-faq/how-many-languages-world-are-unwritten-0} {{How many languages in the world are unwritten?}}

\bibitem[{Galvez et~al.(2021)Galvez, Diamos, Torres, Cer{\'o}n, Achorn, Gopi, Kanter, Lam, Mazumder, and Reddi}]{galvez2021people}
Daniel Galvez, Greg Diamos, Juan Manuel~Ciro Torres, Juan~Felipe Cer{\'o}n, Keith Achorn, Anjali Gopi, David Kanter, Max Lam, Mark Mazumder, and Vijay~Janapa Reddi. 2021.
\newblock \href {https://openreview.net/forum?id=R8CwidgJ0yT} {The {P}eople{\textquoteright}s {S}peech: A large-scale diverse {E}nglish speech recognition dataset for commercial usage}.
\newblock In \emph{NeurIPS 2021}.

\bibitem[{Godfrey et~al.(1992)}]{swbd-corpus}
J.J. Godfrey et~al. 1992.
\newblock {SWITCHBOARD: telephone speech corpus for research and development}.
\newblock In \emph{ICASSP 1992}.

\bibitem[{Graves et~al.(2006)Graves, Fern{\'a}ndez, Gomez, and Schmidhuber}]{graves2006connectionist}
Alex Graves, Santiago Fern{\'a}ndez, Faustino Gomez, and J{\"u}rgen Schmidhuber. 2006.
\newblock Connectionist temporal classification: labelling unsegmented sequence data with recurrent neural networks.
\newblock In \emph{ICML 2006}, pages 369--376.

\bibitem[{Groeneveld et~al.(2024)Groeneveld, Beltagy, Walsh, Bhagia, Kinney, Tafjord, Jha, Ivison, Magnusson, Wang et~al.}]{groeneveld2024olmo}
Dirk Groeneveld, Iz~Beltagy, Pete Walsh, Akshita Bhagia, Rodney Kinney, Oyvind Tafjord, Ananya~Harsh Jha, Hamish Ivison, Ian Magnusson, Yizhong Wang, et~al. 2024.
\newblock Olmo: Accelerating the science of language models.
\newblock \emph{arXiv preprint arXiv:2402.00838}.

\bibitem[{Gulati et~al.(2020)Gulati, Qin, Chiu, Parmar, Zhang, Yu, Han, Wang, Zhang, Wu, and Pang}]{conformer}
Anmol Gulati, James Qin, Chung-Cheng Chiu, Niki Parmar, Yu~Zhang, Jiahui Yu, Wei Han, Shibo Wang, Zhengdong Zhang, Yonghui Wu, and Ruoming Pang. 2020.
\newblock Conformer: Convolution-augmented transformer for speech recognition.
\newblock In \emph{Interspeech 2020}.

\bibitem[{Helgadóttir et~al.(2019)Helgadóttir, Nikulásdóttir, Borský, Fong, Kjaran, and Guðnason}]{helgadottir19_althingi}
Inga~R. Helgadóttir, Anna~Björk Nikulásdóttir, Michal Borský, Judy~Y. Fong, Róbert Kjaran, and Jón Guðnason. 2019.
\newblock \href {https://doi.org/10.21437/Interspeech.2019-1248} {The {A}lthingi {ASR} system}.
\newblock In \emph{Interspeech 2019}.

\bibitem[{Hernandez et~al.(2018)Hernandez, Nguyen, Ghannay, Tomashenko, and Esteve}]{tedlium3}
Fran{\c{c}}ois Hernandez, Vincent Nguyen, Sahar Ghannay, Natalia Tomashenko, and Yannick Esteve. 2018.
\newblock {TED-LIUM} 3: Twice as much data and corpus repartition for experiments on speaker adaptation.
\newblock In \emph{Speech and Computer: 20th International Conference, SPECOM 2018}. Springer.

\bibitem[{Hsu et~al.(2021)Hsu, Bolte, Tsai, Lakhotia, Salakhutdinov, and Mohamed}]{hsuHubert}
Wei-Ning Hsu, Benjamin Bolte, Yao-Hung~Hubert Tsai, Kushal Lakhotia, Ruslan Salakhutdinov, and Abdelrahman Mohamed. 2021.
\newblock \href {https://doi.org/10.1109/TASLP.2021.3122291} {{HuBERT}: Self-supervised speech representation learning by masked prediction of hidden units}.
\newblock \emph{IEEE/ACM TALSP}.

\bibitem[{Huang et~al.(2023)Huang, Fu, Hsu, Gutierrez, Wang, Tseng, Zhang, and Lee}]{distilldistort}
Kuan-Po Huang, Yu-Kuan Fu, Tsu-Yuan Hsu, Fabian~Ritter Gutierrez, Fan-Lin Wang, Liang-Hsuan Tseng, Yu~Zhang, and Hung-yi Lee. 2023.
\newblock \href {https://doi.org/10.1109/SLT54892.2023.10022474} {Improving generalizability of distilled self-supervised speech processing models under distorted settings}.
\newblock In \emph{SLT 2022}, pages 1112--1119.

\bibitem[{{IARPA}()}]{babel}
{IARPA}.
\newblock \href {www.iarpa.gov/index.php/research-programs/babel} {{The Babel Program}}.

\bibitem[{Kahn et~al.(2020)Kahn, Rivière, Zheng, Kharitonov, Xu, Mazaré, Karadayi, Liptchinsky, Collobert, Fuegen, Likhomanenko, Synnaeve, Joulin, Mohamed, and Dupoux}]{kahnLibriLight}
J.~Kahn, M.~Rivière, W.~Zheng, E.~Kharitonov, Q.~Xu, P.E. Mazaré, J.~Karadayi, V.~Liptchinsky, R.~Collobert, C.~Fuegen, T.~Likhomanenko, G.~Synnaeve, A.~Joulin, A.~Mohamed, and E.~Dupoux. 2020.
\newblock \href {https://doi.org/10.1109/ICASSP40776.2020.9052942} {{Libri-Light}: A benchmark for {ASR} with limited or no supervision}.
\newblock In \emph{ICASSP 2020}.

\bibitem[{Kim et~al.(2023)Kim, Wu, Peng, Pan, Sridhar, Han, and Watanabe}]{ebf}
Kwangyoun Kim, Felix Wu, Yifan Peng, Jing Pan, Prashant Sridhar, Kyu~J Han, and Shinji Watanabe. 2023.
\newblock E-branchformer: Branchformer with enhanced merging for speech recognition.
\newblock In \emph{SLT 2023}.

\bibitem[{Kingma and Ba(2015)}]{adam}
Diederik~P Kingma and Jimmy Ba. 2015.
\newblock Adam: A method for stochastic optimization.
\newblock \emph{ICLR 2015}.

\bibitem[{Ko et~al.(2017)Ko, Peddinti, Povey, Seltzer, and Khudanpur}]{rir_noises}
Tom Ko, Vijayaditya Peddinti, Daniel Povey, Michael~L. Seltzer, and Sanjeev Khudanpur. 2017.
\newblock A study on data augmentation of reverberant speech for robust speech recognition.
\newblock In \emph{ICASSP 2017}.

\bibitem[{Kong et~al.(2020)Kong, Kim, and Bae}]{hifigan}
Jungil Kong, Jaehyeon Kim, and Jaekyoung Bae. 2020.
\newblock {Hifi-GAN: Generative Adversarial Networks for Efficient and High Fidelity Speech Synthesis}.
\newblock \emph{NeurIPS 2020}, 33:17022--17033.

\bibitem[{Li et~al.(2022)Li, Metze, Mortensen, Black, and Watanabe}]{asr2k}
Xinjian Li, Florian Metze, David~R. Mortensen, Alan~W Black, and Shinji Watanabe. 2022.
\newblock \href {https://doi.org/10.21437/Interspeech.2022-10712} {{ASR2K: Speech Recognition for Around 2000 Languages without Audio}}.
\newblock In \emph{Interspeech 2022}.

\bibitem[{Li et~al.(2023)Li, Takamichi, Saeki, Chen, Shiota, and Watanabe}]{yodas}
Xinjian Li, Shinnosuke Takamichi, Takaaki Saeki, William Chen, Sayaka Shiota, and Shinji Watanabe. 2023.
\newblock {YODAS}: {Y}outube-oriented dataset for audio and speech.
\newblock In \emph{ASRU 2023}.

\bibitem[{Liu et~al.(2024)Liu, Chang, Auli, Hsu, and Glass}]{liu2024dinosr}
Alexander~H Liu, Heng-Jui Chang, Michael Auli, Wei-Ning Hsu, and Jim Glass. 2024.
\newblock {DinoSR}: Self-distillation and online clustering for self-supervised speech representation learning.
\newblock \emph{NeurIPS 2024}, 36.

\bibitem[{Liu et~al.(2023)Liu, Qiao, Neiswanger, Wang, Tan, Tao, Li, Wang, Sun, Pangarkar et~al.}]{liu2023llm360}
Zhengzhong Liu, Aurick Qiao, Willie Neiswanger, Hongyi Wang, Bowen Tan, Tianhua Tao, Junbo Li, Yuqi Wang, Suqi Sun, Omkar Pangarkar, et~al. 2023.
\newblock {LLM}360: Towards fully transparent open-source llms.
\newblock \emph{arXiv preprint arXiv:2312.06550}.

\bibitem[{Lo et~al.(2019)Lo, Fu, Huang, Wang, Yamagishi, Tsao, and Wang}]{mosnet}
Chen-Chou Lo, Szu-Wei Fu, Wen-Chin Huang, Xin Wang, Junichi Yamagishi, Yu~Tsao, and Hsin-Min Wang. 2019.
\newblock \href {https://doi.org/10.21437/Interspeech.2019-2003} {{MOSNet: Deep Learning-Based Objective Assessment for Voice Conversion}}.
\newblock In \emph{Interspeech 2019}, pages 1541--1545.

\bibitem[{Lyu et~al.(2010)Lyu, Tan, Chng, and Li}]{lyu10_seame}
Dau-Cheng Lyu, Tien-Ping Tan, Eng~Siong Chng, and Haizhou Li. 2010.
\newblock \href {https://doi.org/10.21437/Interspeech.2010-563} {{SEAME}: a {M}andarin-{E}nglish code-switching speech corpus in south-east {A}sia}.
\newblock In \emph{Interspeech 2010}.

\bibitem[{Maiti et~al.(2024)Maiti, Peng, Choi, Jung, Chang, and Watanabe}]{maiti2024voxtlm}
Soumi Maiti, Yifan Peng, Shukjae Choi, Jee-weon Jung, Xuankai Chang, and Shinji Watanabe. 2024.
\newblock {VoxtLM: Unified Decoder-Only Models for Consolidating Speech Recognition, Synthesis and Speech, Text Continuation Tasks}.
\newblock In \emph{ICASSP 2024}, pages 13326--13330. IEEE.

\bibitem[{Ng et~al.(2023)Ng, Zhang, Yip, Yang, Ni, Zhang, Ma, Ni, Chng, and Ma}]{dehubert}
Dianwen Ng, Ruixi Zhang, Jia~Qi Yip, Zhao Yang, Jinjie Ni, Chong Zhang, Yukun Ma, Chongjia Ni, Eng~Siong Chng, and Bin Ma. 2023.
\newblock \href {https://doi.org/10.1109/ICASSP49357.2023.10096603} {De’hubert: Disentangling noise in a self-supervised model for robust speech recognition}.
\newblock In \emph{ICASSP 2023}, pages 1--5.

\bibitem[{Nozaki and Komatsu(2021)}]{nozaki2021relaxing}
Jumon Nozaki and Tatsuya Komatsu. 2021.
\newblock Relaxing the conditional independence assumption of {CTC}-based {ASR} by conditioning on intermediate predictions.
\newblock In \emph{Interspeech 2021}, pages 3735--3739.

\bibitem[{O’Neill et~al.(2021)O’Neill, Lavrukhin, Majumdar, Noroozi, Zhang, Kuchaiev, Balam, Dovzhenko, Freyberg, Shulman, Ginsburg, Watanabe, and Kucsko}]{spgispeech}
Patrick~K. O’Neill, Vitaly Lavrukhin, Somshubra Majumdar, Vahid Noroozi, Yuekai Zhang, Oleksii Kuchaiev, Jagadeesh Balam, Yuliya Dovzhenko, Keenan Freyberg, Michael~D. Shulman, Boris Ginsburg, Shinji Watanabe, and Georg Kucsko. 2021.
\newblock {SPGISpeech}: 5,000 hours of transcribed financial audio for fully formatted end-to-end speech recognition.
\newblock In \emph{Interspeech 2021}.

\bibitem[{Panayotov et~al.(2015)}]{librispeech-corpus}
Vassil Panayotov et~al. 2015.
\newblock Librispeech: An {ASR} corpus based on public domain audio books.
\newblock In \emph{ICASSP 2015}.

\bibitem[{Paul and Baker(1992)}]{wsj}
Douglas~B Paul and Janet Baker. 1992.
\newblock {The design for the Wall Street Journal-based CSR corpus}.
\newblock In \emph{Workshop on Speech and Natural Language}.

\bibitem[{Peng et~al.(2022)Peng, Dalmia, Lane, and Watanabe}]{branchformer}
Yifan Peng, Siddharth Dalmia, Ian Lane, and Shinji Watanabe. 2022.
\newblock Branchformer: Parallel {MLP}-attention architectures to capture local and global context for speech recognition and understanding.
\newblock In \emph{ICML 2022}.

\bibitem[{Peng et~al.(2023{\natexlab{a}})Peng, Kim, Wu, Yan, Arora, Chen, Tang, Shon, Sridhar, and Watanabe}]{ebf-vs-conformer}
Yifan Peng, Kwangyoun Kim, Felix Wu, Brian Yan, Siddhant Arora, William Chen, Jiyang Tang, Suwon Shon, Prashant Sridhar, and Shinji Watanabe. 2023{\natexlab{a}}.
\newblock A comparative study on e-branchformer vs conformer in speech recognition, translation, and understanding tasks.
\newblock In \emph{Interspeech 2023}.

\bibitem[{Peng et~al.(2024{\natexlab{a}})Peng, Sudo, Shakeel, and Watanabe}]{peng2024owsmctc}
Yifan Peng, Yui Sudo, Muhammad Shakeel, and Shinji Watanabe. 2024{\natexlab{a}}.
\newblock {OWSM-CTC: An Open Encoder-Only Speech Foundation Model for Speech Recognition, Translation, and Language Identification}.
\newblock \emph{arXiv preprint arXiv:2402.12654}.

\bibitem[{Peng et~al.(2024{\natexlab{b}})Peng, Tian, Chen, Arora, Yan, Sudo, Shakeel, Choi, Shi, Chang et~al.}]{peng2024owsm}
Yifan Peng, Jinchuan Tian, William Chen, Siddhant Arora, Brian Yan, Yui Sudo, Muhammad Shakeel, Kwanghee Choi, Jiatong Shi, Xuankai Chang, et~al. 2024{\natexlab{b}}.
\newblock {OWSM v3. 1: Better and Faster Open Whisper-Style Speech Models based on E-Branchformer}.
\newblock \emph{arXiv preprint arXiv:2401.16658}.

\bibitem[{Peng et~al.(2023{\natexlab{b}})Peng, Tian, Yan, Berrebbi, Chang, Li, Shi, Arora, Chen, Sharma, Zhang, Sudo, Shakeel, weon Jung, Maiti, and Watanabe}]{asru23-owsm}
Yifan Peng, Jinchuan Tian, Brian Yan, Dan Berrebbi, Xuankai Chang, Xinjian Li, Jiatong Shi, Siddhant Arora, William Chen, Roshan Sharma, Wangyou Zhang, Yui Sudo, Muhammad Shakeel, Jee weon Jung, Soumi Maiti, and Shinji Watanabe. 2023{\natexlab{b}}.
\newblock Reproducing {W}hisper-style training using an open-source toolkit and publicly available data.
\newblock In \emph{ASRU 2023}.

\bibitem[{Polyak et~al.(2021)Polyak, Adi, Copet, Kharitonov, Lakhotia, Hsu, Mohamed, and Dupoux}]{resynthesis}
Adam Polyak, Yossi Adi, Jade Copet, Eugene Kharitonov, Kushal Lakhotia, Wei-Ning Hsu, Abdelrahman Mohamed, and Emmanuel Dupoux. 2021.
\newblock {Speech Resynthesis from Discrete Disentangled Self-Supervised Representations}.
\newblock In \emph{Interspeech 2021}.

\bibitem[{Post et~al.(2013)}]{fisher-callhome}
Matt Post et~al. 2013.
\newblock Improved speech-to-text translation with the fisher and callhome {S}panish-{E}nglish speech translation corpus.
\newblock In \emph{IWSLT 2013}.

\bibitem[{Pratap et~al.(2023)Pratap, Tjandra, Shi, Tomasello, Babu, Kundu, Elkahky, Ni, Vyas, Fazel-Zarandi et~al.}]{pratap2023scaling}
Vineel Pratap, Andros Tjandra, Bowen Shi, Paden Tomasello, Arun Babu, Sayani Kundu, Ali Elkahky, Zhaoheng Ni, Apoorv Vyas, Maryam Fazel-Zarandi, et~al. 2023.
\newblock Scaling speech technology to 1,000+ languages.
\newblock \emph{arxiv:2305.13516}.

\bibitem[{Pratap et~al.()Pratap, Xu, Sriram, Synnaeve, and Collobert}]{pratap2020mls}
Vineel Pratap, Qiantong Xu, Anuroop Sriram, Gabriel Synnaeve, and Ronan Collobert.
\newblock {MLS}: A large-scale multilingual dataset for speech research.
\newblock In \emph{Interspeech 2020}, pages 2757--2761.

\bibitem[{Radford et~al.(2023)Radford, Kim, Xu, Brockman, Mcleavey, and Sutskever}]{whisper}
Alec Radford, Jong~Wook Kim, Tao Xu, Greg Brockman, Christine Mcleavey, and Ilya Sutskever. 2023.
\newblock Robust speech recognition via large-scale weak supervision.
\newblock In \emph{ICML 2023}.

\bibitem[{Ravanelli et~al.(2021)Ravanelli, Parcollet, Plantinga, Rouhe, Cornell, Lugosch, Subakan, Dawalatabad, Heba, Zhong et~al.}]{ravanelli2021speechbrain}
Mirco Ravanelli, Titouan Parcollet, Peter Plantinga, Aku Rouhe, Samuele Cornell, Loren Lugosch, Cem Subakan, Nauman Dawalatabad, Abdelwahab Heba, Jianyuan Zhong, et~al. 2021.
\newblock Speechbrain: A general-purpose speech toolkit.
\newblock \emph{arXiv preprint arXiv:2106.04624}.

\bibitem[{Reddy et~al.(2021)Reddy, Dubey, Koishida, Nair, Gopal, Cutler, Braun, Gamper, Aichner, and Srinivasan}]{dns_noise}
Chandan~K.A. Reddy, Harishchandra Dubey, Kazuhito Koishida, Arun Nair, Vishak Gopal, Ross Cutler, Sebastian Braun, Hannes Gamper, Robert Aichner, and Sriram Srinivasan. 2021.
\newblock {INTERSPEECH 2021 Deep Noise Suppression Challenge}.
\newblock In \emph{Interspeech 2021}.

\bibitem[{Salesky et~al.(2021)Salesky, Wiesner, Bremerman, Cattoni, Negri, Turchi, Oard, and Post}]{salesky2021mtedx}
Elizabeth Salesky, Matthew Wiesner, Jacob Bremerman, Roldano Cattoni, Matteo Negri, Marco Turchi, Douglas~W. Oard, and Matt Post. 2021.
\newblock Multilingual {TEDx} corpus for speech recognition and translation.
\newblock In \emph{Interspeech 2021}.

\bibitem[{Sanabria et~al.(2023)Sanabria, Bogoychev, Markl, Carmantini, Klejch, and Bell}]{edinburgh}
Ramon Sanabria, Nikolay Bogoychev, Nina Markl, Andrea Carmantini, Ondrej Klejch, and Peter Bell. 2023.
\newblock The {Edinburgh} international accents of {English} corpus: Towards the democratization of {English} {ASR}.
\newblock In \emph{ICASSP 2023}.

\bibitem[{Shi et~al.(2021{\natexlab{a}})Shi, Amith, Chang, Dalmia, Yan, and Watanabe}]{shi2021highland}
Jiatong Shi, Jonathan~D Amith, Xuankai Chang, Siddharth Dalmia, Brian Yan, and Shinji Watanabe. 2021{\natexlab{a}}.
\newblock Highland {P}uebla {N}ahuatl speech translation corpus for endangered language documentation.
\newblock In \emph{AmericasNLP 2021}.

\bibitem[{Shi et~al.(2021{\natexlab{b}})Shi, Amith, Garc{\'\i}a, Sierra, Duh, and Watanabe}]{shi2021mixtec}
Jiatong Shi, Jonathan~D Amith, Rey~Castillo Garc{\'\i}a, Esteban~Guadalupe Sierra, Kevin Duh, and Shinji Watanabe. 2021{\natexlab{b}}.
\newblock Leveraging end-to-end {ASR} for endangered language documentation: An empirical study on {Y}ol{\'o}xochitl {M}ixtec.
\newblock In \emph{EACL 2021}.

\bibitem[{Shi et~al.(2023{\natexlab{a}})Shi, Berrebbi, Chen, Hu, Huang, Chung, Chang, Li, Mohamed, yi~Lee, and Watanabe}]{mlsuperb}
Jiatong Shi, Dan Berrebbi, William Chen, En-Pei Hu, Wei-Ping Huang, Ho-Lam Chung, Xuankai Chang, Shang-Wen Li, Abdelrahman Mohamed, Hung yi~Lee, and Shinji Watanabe. 2023{\natexlab{a}}.
\newblock \href {https://doi.org/10.21437/Interspeech.2023-1316} {{ML-SUPERB: Multilingual Speech Universal PERformance Benchmark}}.
\newblock In \emph{Interspeech 2023}.

\bibitem[{Shi et~al.(2023{\natexlab{b}})Shi, Chen, Berrebbi, Wang, Huang, Hu, Chuang, Chang, Tang, Li, Mohamed, Lee, and Watanabe}]{mlsuperbchallenge}
Jiatong Shi, William Chen, Dan Berrebbi, Hsiu-Hsuan Wang, Wei-Ping Huang, En-Pei Hu, Ho-Lam Chuang, Xuankai Chang, Yuxun Tang, Shang-Wen Li, Abdelrahman Mohamed, Hung-Yi Lee, and Shinji Watanabe. 2023{\natexlab{b}}.
\newblock Findings of the 2023 {ML-SUPERB} challenge: Pre-training and evaluation over more languages and beyond.
\newblock In \emph{ASRu 2023}.

\bibitem[{Shi et~al.(2024)Shi, Inaguma, Ma, Kulikov, and Sun}]{shi2024multiresolution}
Jiatong Shi, Hirofumi Inaguma, Xutai Ma, Ilia Kulikov, and Anna Sun. 2024.
\newblock Multi-resolution {H}u{BERT}: Multi-resolution speech self-supervised learning with masked unit prediction.
\newblock In \emph{ICLR 2024}.

\bibitem[{Slizhikova et~al.(2020)}]{ru-open-stt}
Anna Slizhikova et~al. 2020.
\newblock \href {https://github.com/snakers4/open_stt} {{Russian Open Speech To Text (STT/ASR) Dataset}}.

\bibitem[{Smule(2019)}]{dampmvp}
Inc Smule. 2019.
\newblock \href {https://doi.org/10.5281/zenodo.2747436} {{DAMP-MVP: Digital Archive of Mobile Performances - Smule Multilingual Vocal Performance 300x30x2}}.

\bibitem[{Solberg and Ortiz(2022)}]{solberg-ortiz-2022-norwegian}
Per~Erik Solberg and Pablo Ortiz. 2022.
\newblock The {N}orwegian parliamentary speech corpus.
\newblock In \emph{LREC 2022}, Marseille, France.

\bibitem[{Valk and Alumäe(2021)}]{voxlingua}
Jörgen Valk and Tanel Alumäe. 2021.
\newblock {VOXLINGUA107}: A dataset for spoken language recognition.
\newblock In \emph{SLT 2021}.

\bibitem[{Vaswani et~al.(2017)Vaswani, Shazeer, Parmar, Uszkoreit, Jones, Gomez, Kaiser, and Polosukhin}]{transformer}
Ashish Vaswani, Noam Shazeer, Niki Parmar, Jakob Uszkoreit, Llion Jones, Aidan~N. Gomez, Lukasz Kaiser, and Illia Polosukhin. 2017.
\newblock Attention is all you need.
\newblock In \emph{NeurIPS 2017}.

\bibitem[{VoxForge()}]{voxforge}
VoxForge.
\newblock \href {http://www.voxforge.org/} {{VoxForge}}.

\bibitem[{Wang et~al.(2020)Wang, Tang, Ma, Wu, Popuri, Okhonko, and Pino}]{wang2020fairseq}
Changhan Wang, Yun Tang, Xutai Ma, Anne Wu, Sravya Popuri, Dmytro Okhonko, and Juan Pino. 2020.
\newblock Fairseq {S2T}: Fast speech-to-text modeling with fairseq.
\newblock \emph{arxiv:2010.05171}.

\bibitem[{Wang et~al.(2021)}]{voxpopuli}
Changhan Wang et~al. 2021.
\newblock {VoxPopuli: A Large-Scale Multilingual Speech Corpus for Representation Learning, Semi-Supervised Learning and Interpretation}.
\newblock In \emph{ACL 2021}.

\bibitem[{Watanabe et~al.(2018)Watanabe, Hori, Karita, Hayashi, Nishitoba, Unno, {Enrique Yalta Soplin}, Heymann, Wiesner, Chen, Renduchintala, and Ochiai}]{espnet}
Shinji Watanabe, Takaaki Hori, Shigeki Karita, Tomoki Hayashi, Jiro Nishitoba, Yuya Unno, Nelson {Enrique Yalta Soplin}, Jahn Heymann, Matthew Wiesner, Nanxin Chen, Adithya Renduchintala, and Tsubasa Ochiai. 2018.
\newblock {ESP}net: End-to-end speech processing toolkit.
\newblock In \emph{Interspeech 2018}.

\bibitem[{Watanabe et~al.(2017)Watanabe, Hori, Kim, Hershey, and Hayashi}]{watanabehybrid}
Shinji Watanabe, Takaaki Hori, Suyoun Kim, John~R. Hershey, and Tomoki Hayashi. 2017.
\newblock Hybrid {CTC}/attention architecture for end-to-end speech recognition.
\newblock \emph{IEEE Journal of Selected Topics in Signal Processing}.

\bibitem[{wen Yang et~al.(2021)wen Yang, Chi, Chuang, Lai, Lakhotia, Lin, Liu, Shi, Chang, Lin, Huang, Tseng, tik Lee, Liu, Huang, Dong, Li, Watanabe, Mohamed, and yi~Lee}]{superb}
Shu wen Yang, Po-Han Chi, Yung-Sung Chuang, Cheng-I~Jeff Lai, Kushal Lakhotia, Yist~Y. Lin, Andy~T. Liu, Jiatong Shi, Xuankai Chang, Guan-Ting Lin, Tzu-Hsien Huang, Wei-Cheng Tseng, Ko~tik Lee, Da-Rong Liu, Zili Huang, Shuyan Dong, Shang-Wen Li, Shinji Watanabe, Abdelrahman Mohamed, and Hung yi~Lee. 2021.
\newblock {SUPERB: Speech Processing Universal PERformance Benchmark}.
\newblock In \emph{Interspeech 2021}.

\bibitem[{Workshop et~al.(2022)Workshop, Scao, Fan, Akiki, Pavlick, Ili{\'c}, Hesslow, Castagn{\'e}, Luccioni, Yvon et~al.}]{workshop2022bloom}
BigScience Workshop, Teven~Le Scao, Angela Fan, Christopher Akiki, Ellie Pavlick, Suzana Ili{\'c}, Daniel Hesslow, Roman Castagn{\'e}, Alexandra~Sasha Luccioni, Fran{\c{c}}ois Yvon, et~al. 2022.
\newblock Bloom: A 176b-parameter open-access multilingual language model.
\newblock \emph{arXiv preprint arXiv:2211.05100}.

\bibitem[{Yamagishi et~al.(2019)}]{vctk}
Junichi Yamagishi et~al. 2019.
\newblock {CSTR VCTK Corpus: English Multi-speaker Corpus for CSTR Voice Cloning Toolkit}.

\bibitem[{Yang et~al.(2023)Yang, Ma, Zheng, Song, Niu, and Chen}]{fasthubert}
Guanrou Yang, Ziyang Ma, Zhisheng Zheng, Yakun Song, Zhikang Niu, and Xie Chen. 2023.
\newblock Fast-{H}u{BERT}: an efficient training framework for self-supervised speech representation learning.
\newblock In \emph{ASRU 2023}.

\bibitem[{Yang et~al.(2022)Yang, Chen, Luo, Yang, Ye, Cheng, Xu, Jin, Zhang, Zhang, Xie, and Yan}]{magicdata}
Zehui Yang, Yifan Chen, Lei Luo, Runyan Yang, Lingxuan Ye, Gaofeng Cheng, Ji~Xu, Yaohui Jin, Qingqing Zhang, Pengyuan Zhang, Lei Xie, and Yonghong Yan. 2022.
\newblock Open source {MagicData-RAMC}: A rich annotated mandarin conversational ({RAMC}) speech dataset.
\newblock In \emph{Interspeech 2022}, pages 1736--1740.

\bibitem[{Yin et~al.(2023)Yin, Mori et~al.}]{reazonspeech}
Yue Yin, Daijiro Mori, et~al. 2023.
\newblock {ReazonSpeech: A Free and Massive Corpus for Japanese ASR}.

\bibitem[{Zhang et~al.(2022)}]{wenetspeech}
Binbin Zhang et~al. 2022.
\newblock Wenet{S}peech: A 10000+ hours multi-domain mandarin corpus for speech recognition.
\newblock In \emph{ICASSP 2022}.

\bibitem[{Zhang et~al.(2023)Zhang, Han, Qin, Wang, Bapna, Chen, Chen, Li, Axelrod, Wang et~al.}]{google-usm}
Yu~Zhang, Wei Han, James Qin, Yongqiang Wang, Ankur Bapna, Zhehuai Chen, Nanxin Chen, Bo~Li, Vera Axelrod, Gary Wang, et~al. 2023.
\newblock Google {USM}: Scaling automatic speech recognition beyond 100 languages.
\newblock \emph{arxiv:2303.01037}.

\bibitem[{Zhu et~al.(2023)Zhu, Zhou, Zhang, Liu, Hu, and Dai}]{robustd2v}
Qiu-Shi Zhu, Long Zhou, Jie Zhang, Shu-Jie Liu, Yu-Chen Hu, and Li-Rong Dai. 2023.
\newblock \href {https://doi.org/10.1109/ICASSP49357.2023.10095373} {Robust data2vec: Noise-robust speech representation learning for asr by combining regression and improved contrastive learning}.
\newblock In \emph{ICASSP 2023}, pages 1--5.

\end{thebibliography}

\clearpage
\appendix
\section{Appendix}
\label{sec:appendix}

\subsection{Data} \label{sec:a_data}
For brevity, we provide an overview of these datasets in Table \ref{tab:data} and refer readers to the original papers for more details \cite{aishell-corpus,gigaspeech,librispeech-corpus,spgispeech,tedlium3,pratap2020mls,wenetspeech,ami-corpus,commonvoice,swbd-corpus,FLEURS,ksponspeech,magicdata,voxpopuli,wavelablm, yodas, kahnLibriLight, galvez2021people, voxlingua, helgadottir19_althingi, lyu10_seame, solberg-ortiz-2022-norwegian, shi2021highland, shi2021mixtec, cardenas2018siminchik, edinburgh, accentdb, berrebbi22_totonac}. 

Over 80\% of the pre-training data is derived from two corpora: YODAS \cite{yodas} and VoxPopuli \cite{voxpopuli}. However, both of these sources largely consist of European languages. To add more linguistic diversity, we also smaller scale multilingual corpora such as include BABEL \cite{babel}, Googlei18n \cite{wavelablm}, VoxLingua \cite{voxlingua}, and FLEURS \cite{FLEURS}. While some may argue that FLEURS and/or BABEL was originally designed to be out-of-domain evaluation, we note that many works now include it as an in-domain dataset for both supervised and unsupervised training \cite{asru23-owsm, pratap2023scaling, wavelablm, google-usm}. 

We complement this selection of multilingual corpora with various language-specific data to boost the representation of languages that are underrepresented in large web-scale datasets. This includes many indigenous languages, such as Quechua \cite{cardenas2018siminchik}, Mixtec \cite{shi2021mixtec}, and Totonac \cite{berrebbi22_totonac}.

Finally, we also make an effort to support speech outside of mainstream accents and voices. For example, we include code-switching \cite{lyu10_seame}, accented data \cite{nptel, edinburgh, accentdb}, and even singing voices \cite{dampmvp}.

\subsection{Ablations} \label{sec:ablation_arch}

To understand the effectiveness of E-Branchformer and WavLM denoising in the multilingual setting, we conducted several SSL experiments at a smaller scale. We sampled 7,000 hours of data from our 1 million hour SSL dataset and trained a 95M parameter HuBERT Base model on that data (ML-HuBERT 7K). Then, we trained a variant with the addition of the acoustic denoising task and another with the Transformer layers replaced with E-Branchformer layers. We then benchmarked each model on ML-SUPERB using the same settings as Section \ref{ml-superb}. Our results show meaningful gains achieved with the addition of the WavLM objective and E-Branchformer architecture.

\begin{table}[htb]
    \centering
    \caption{Ablations on the ML-SUPERB 1-hour set.}
    \resizebox{\columnwidth}{!}{
    \begin{tabular}{l|ccc}
    \toprule
     Model  & \multicolumn{3}{c}{Multi.ASR + LID}\\
     &  CER & CER (Few-Shot) & ACC \\
     \midrule
     ML-HuBERT 7K & 36.9 & \textbf{40.5} & 78.5 \\
     + Denoising & 30.5 & 41.2 & 84.3 \\
     + E-Branchformer & \textbf{29.0} & 40.9 & \textbf{85.2} \\
     \bottomrule
    \end{tabular}}
    \label{tab:ablation-ml}
\end{table}

\subsection{Dereverberation} \label{sec:dereverb-ablation}
We perform a preliminary investigation of on the effectiveness of our proposed SSL dereverberation task (Section \ref{sec:dereverb}) by training two HuBERT Base models \cite{hsuHubert} on LibriSpeech 960 \cite{librispeech-corpus}. The first model corresponds to the vanilla HuBERT setting without any form of data augmentation, while the second model is trained with our proposed dereverberation technique. We fine-tune each model on 10-hours of LibriLight ASR \cite{kahnLibriLight}, where we use performance on test-clean and test-other as a proxy for comparing performance on clean and noisy data respectively. Results are in Table \ref{tab:librilight}, with improvements across both test sets. The benefits are more significant on the noisier test-other, with relative reductions of 6.9\% Word Error Rate (WER), indicating the effectiveness of our technique for enhancing model robustness. Performance on test-clean is also improved by 3.1\%.

\begin{table}[h]
    \centering
    \caption{Investigation on effectiveness of dereverberation augmentation on English ASR by WER.}
    \vspace{-0.2cm}
    \begin{tabular}{l|c c}
    \toprule
    Model &  test-clean & test-other\\
    \midrule
    HuBERT & 13.1 & 22.6 \\
    + Dereverberation & \textbf{12.7} & \textbf{21.1}\\
    \bottomrule
    \end{tabular}
    \label{tab:librilight}
\end{table}

\subsection{Pre-Training Details} \label{a_pt}

\begin{table}[h]
    \centering
    \caption{Computation budget of XEUS in CPU/GPU hours. Reported numbers for formatting are in CPU hours, while all other stages are measured in GPU hours.}
    \begin{tabular}{l|r}
    \toprule
    Stage & Hours \\
    \midrule
    Data Preparation & 100,000 \\
    Pseudo-labelling & 15,000 \\
    Ablations & 2,100 \\
    Scaling & 200 \\
    Pre-Training & 63,000 \\
    \bottomrule
    \end{tabular}
    \label{tab:costs}
\end{table}

Table \ref{tab:costs} provides a breakdown of XEUS' the computational costs, measured in CPU and GPU hours. We used AMD EPYC 7763 processors for CPU-based jobs. For most GPU tasks, we use a mix of Nvidia A40 46GB and Nvidia A100 40GB GPUs. For pre-training the final model, we exclusively used A100s. 

We consumed around 100K CPU hours for the data stage. The bulk of this usage came from processing the YODAS dataset, which originally consisted of document-level WAV files at various sampling rates. We had to convert each file into 16kHz audio samples and then segment each into utterance-level pieces. Another large source of CPU consumption came from downloading each dataset and unarchiving them onto the disk, which became a non-trivial effort for larger datasets such as VoxPopuli.

Obtaining the HuBERT pseudo-labels required a large amount of compute, totalling 15,000 GPU hours. While one may argue that this process could have been avoid by using another SSL objective, such as data2vec \cite{baevski2022data2vec} or w2v-BERT \cite{chungW2vBert}, we believed that this cost was worth the guaranteed stability during large-scale training. Self-distilling SSL objectives like data2vec are prone to representation collapse \cite{baevski2022data2vec, liu2024dinosr}, while wav2vec derived models require a codebook diversity loss \cite{baevskiw2v, chungW2vBert}. As our experimental runs were limited, we believed the simple HuBERT objective would be the easiest to scale.

The ablations described in Sections \ref{sec:dereverb-ablation} and \ref{sec:ablation_arch} consumed a total of 2,100 GPU hours, while the hyperparameter tuning required for scaling (masking ratio, learning rate, warmup steps) consumed 200 GPU hours. 

The largest source of compute usage came from the final pre-training phase, which consumed 63,000 GPU hours across a total of 40 days. We use bfloat16 mixed precision with Flash Attention v2 \cite{flashattn} for faster pre-training. To improve convergence, for the first 3,000 steps we include an additional intermediate cross entropy SSL loss \cite{fasthubert}, an initial masking probability of 0.65, and no data augmentation. This intermediate loss is applied to the 10th encoder layer and is weighted by a factor of 0.3. Afterwards, we remove the intermediate loss for greater efficiency, increase the masking probability to 0.8, and enable the above augmentation methods. We use the Adam optimizer with 32,000 warmup steps and a peak learning rate of 0.0003. 

\subsection{Engineering Details} \label{a_eng}

This section expands upon the engineering optimizations that we made on the ESPnet toolkit mentioned in Section \ref{sec:pretraining}. We organize it into two main components: GPU communication and batching. Quantitative analyses of these optimizations are reported in Table \ref{tab:engineering}.

\begin{table}[h]
    \centering
    \caption{Quantitative results of our engineering improvements. We report percent increases in relative throughput  and percent decreases in relative memory usage after our optimizations.}
    \label{tab:engineering}
    \resizebox {\columnwidth} {!} {
    \begin{tabular}{l|cc}
    \toprule
     Optimization & Throughput ($\uparrow$) & Memory ($\downarrow$) \\
     \midrule
     GPU Synchronization    & 120\% & - \\
    Batch Optimization & 60\% & 113\% \\
     \bottomrule
    \end{tabular}}
\end{table}

\noindent \textbf{GPU Communication: } As the dataset size increases, scaling to a larger number of GPUs is important to finish pre-training in a reasonable time. However, this leads to non-trivial overhead due to communication costs across multiple compute nodes. During our training, we found and removed 2 unnecessary GPU synchronization steps within the ESPnet training code. We also disabled synchronization in iterations without an optimization step (when using gradient accumulation), further improving runtime. These changes are particularly impactful in compute clusters that lack Infiniband support for inter-node communication, which is common in more academic-oriented data centers.

\noindent \textbf{Batching: } The batching method has a large impact on both the memory efficiency and throughput of model training. Ideally, utterances of similar length should be batched together to reduce the amount of padding. This can be complicated during multi-node training, especially when there is large amounts of variance between utterance lengths. We found that ESPnet had issues over-allocating data to batches, which had remained hidden as it only noticeable given a sufficiently large batch size and number of GPUs (such as our case). Fixing this issue more than halved our memory usage given a fixed batch size. We also found that ESPnet randomizes the batches across GPUs regardless of the sequence length. This means that one GPU may process a few utterances that are 30 seconds long, while another may process many utterances less than a second long. As the GPUs need to be synchronized during the backwards pass, this sequence length mismatch causes unnecessary waiting. By enforcing length-aware batch distribution, we are able to improve the model throughput by 60\%.

\subsection{Experimental Setups} \label{a_exp}
This section details the hyperparameters we searched for our experiments, which we then used to obtain and report the best performing results for each model.

\subsubsection{ML-SUPERB: } \label{sec:a_mlsuperb}
The ML-SUPERB benchmark is designed to be a lightweight downstream probe of the multilingual representation quality of SSL models. The benchmark is split across two data settings: 10 minutes and 1 hour of data for each language. Each data setting has 4 tasks: monolingual ASR across 9 languages, multilingual ASR, LID, and joint multilingual ASR+LID. In the multilingual ASR tasks, 5 languages are reserved as few-shot task with 5 examples per language, while the remaining 138 languages have the standard 10 min. / 1 hour of fine-tuning data. An overall SUPERB$_{s}$ score for each model $u$ is calculated relative to the performance of the best performing model for each task $t$ and filterbank-based features. This is obtained with the following formula:
\begin{equation}
    \begin{aligned}[b]
     & \text{SUPERB}_{s}(u) = \\ 
    & \frac{1000}{T} \sum^{T}_{t} \frac{1}{M_t} \sum^{M_t}_{m} 
                \frac{s_{t,m}(u) - s_{t,m}(\text{filterbank})}{s_{t,m}(\text{SOTA}) - s_{t,m}(\text{filterbank})}
\end{aligned}
\end{equation}
Where $M_t$ is the set of metrics for task $t$, such that $s_{t,m}(u)$ yields the score of model $u$ on metric $m$ in $M_t$ for task $t$. SOTA represents the best model for a given metric of each task. Since our model sets new multiple SOTA results, we re-calculate this score for each model following \citet{mlsuperb, mlsuperbchallenge}.

The downstream probe is a 2-layer Transformer encoder trained using CTC \cite{graves2006connectionist} loss. It has a hidden size of 256, a feed forward size of 1024, and 8 attention heads. Each task has a fixed number of training steps and enforces a constant learning scheduler with the Adam \cite{adam} optimizer. The only hyperparameter we adjust during the evaluation is the learning rate, testing values of 0.00004, 0.0001, and 0.0004.

\begin{table*}[tb]
    \centering
    \caption{Overview of datasets used for pre-training. The language column indicates the language used in monolingual datasets and the number of languages in multilingual datasets. \textbf{Bolded} dataset names indicate new corpora we will release.  }
    \resizebox{\textwidth}{!}{
    \begin{tabular}{l|c|c|c|r}
    \toprule
    Dataset  &  License & Language(s) & Domain & Hours\\
    \midrule
    YODAS  \cite{yodas}  & CC BY 3.0 & 140 & Variety & 422K \\
    VoxPopuli \cite{voxpopuli} & CC BY-NC 4.0 & 23 & Legal & 400K \\
    LibriLight \cite{kahnLibriLight} & MIT & English & Audiobook & 60K \\
    MLS \cite{pratap2020mls} & CC BY 4.0 &  8 & Audiobook & 44K \\
    People's Speech \cite{galvez2021people} & CC-BY 4.0 & English & Variety & 30K \\
    WeNetSpeech \cite{wenetspeech} & CC BY 4.0/SA & Mandarin & Variety & 22K \\
    {Russian Open STT} \cite{ru-open-stt} & CC-BY-NC & Russian & Variety & 20K \\
    {NPTEL2020} \cite{nptel} & CC & Indian English & Talk & 15K \\
    {Reazonspeech} \cite{reazonspeech} & Apache 2.0 & Japanese & Television & 15K \\
    Common Voice 13 \cite{commonvoice} & CC0-1.0 & 92 & Read & 13K \\
    GigaSpeech \cite{gigaspeech} & Apache 2.0 & English & Variety & 10K \\
    VoxLingua  \cite{voxlingua} & CC BY 4.0 & 107 & Variety & 6800 \\
    \textbf{MMS-unlab v2} & CC BY-NC 4.0 & 4,023 & Religious & 6700 \\
    SPGI \cite{spgispeech} & - & English & Finance & 5000 \\
    Fisher \cite{fisher-callhome} & LDC & English & Conversation & 2000 \\
    Googlei18n \cite{wavelablm} & Varies & 34 & Variety & 1328\\
    {BABEL} \cite{babel} & IARPA Babel License & 17 & Conversation & 1000 \\
    FLEURS \cite{FLEURS} & CC BY 4.0 & 102 & News & 1000 \\
    KSponSpeech \cite{ksponspeech} & MIT & Korean & Conversation & 970 \\
    LibriSpeech \cite{librispeech-corpus} & CC BY 4.0 & English & Audiobook & 960 \\
    MagicData \cite{magicdata} & CC BY-NC-ND 4.0 & Mandarin & Conversation & 755 \\
    mTEDx \cite{salesky2021mtedx} & CC BY-NC-ND 4.0 & 8 & Talk & 753 \\
    \textbf{Jesus Dramas} & CC BY-NC 4.0 & 430 & Religious & 643 \\
    Althingi \cite{helgadottir19_althingi} & Apache 2.0 & Icelandic & Legal & 542 \\
    TEDLIUM3 \cite{tedlium3} & CC BY-NC-ND 3.0 & English & Talk & 500 \\
    {VoxForge} \cite{voxforge} & GPL & 8 & Read & 235 \\
    AISHELL \cite{aishell-corpus} & Apache 2.0 & Mandarin & Read & 200 \\
    SEAME  \cite{lyu10_seame} & LDC & Codeswitching & Conversation & 192 \\
    {DAMP-MVP} \cite{dampmvp} & Smule Research Data License & English & Singing & 150 \\
    NorwegianParl. \cite{solberg-ortiz-2022-norwegian} & CC0 & Norwegian & Legal & 140 \\
    {AIDATATANG} \cite{aidatatang} & CC BY-NC-ND 4.0 & Mandarin & Read & 140 \\
    AMI \cite{ami-corpus} & CC BY 4.0 & English & Meetings & 100 \\
    Nahuatl \cite{shi2021highland} & CC BY-NC-SA 3.0 & Nahuatl & Conversation & 82 \\
    WSJ \cite{wsj} & LDC & English & Read & 81 \\
    Mixtec \cite{shi2021mixtec} & CC BY-NC-SA 3.0 & Mixtec & Conversation & 70 \\
    \textbf{WikiTongues} & CC BY-NC 4.0 & 700 & Conversation & 70 \\
    Siminchik \cite{cardenas2018siminchik} & CC BY-NC-ND 3.0 &  Quechua & Radio & 50 \\ 
    Edinburgh Accent \cite{edinburgh} & CC-BY-SA & Accented English & Conversation & 40 \\
    {VCTK} \cite{vctk}  & CC BY 4.0 & English & Read & 25 \\
    AccentDB \cite{accentdb} & CC BY-NC 4.0 & Indian English & Read & 20 \\
    Totonac \cite{berrebbi22_totonac} & CC BY-NC-SA 3.0 & Totonac & Monologue & 17 \\
    \bottomrule
    \end{tabular}}
    \label{tab:data}
\end{table*}

\subsubsection{FLEURS: } \label{sec:a_fleurs}
We adopt an identical setup to \cite{ebf-vs-conformer} for the FLEURS evaluations, which remains the SOTA design when not using additional labeled data. The downstream model  consists of a 12 layer E-Branchformer encoder paired with a 6 layer Transformer decoder. The SSL model remains frozen, and a weighted sum of its layer-wise inputs are input into the downstream encoder. Each encoder layer has a convolution kernel of 31, 8 attention heads, a hidden size of 512, and a feed forward size of 2048. Each decoder layer also has 8 attention heads, a hidden size of 512, and a feed forward size of 2048. The model is trained with the joint CTC/attention \cite{watanabehybrid} objective with a CTC weight of 0.3. To better model the diversity in language, the encoder is trained with self-conditioned CTC \cite{nozaki2021relaxing, chenImproving} using the LID+ASR labels. Inference is performed using joint CTC/attention decoding with a language model, using a beam size of 10, a CTC weight of 0.3, and language model weight of 0.4. The model is trained using the Adam optimizer with a Noam-style learning rate scheduler \cite{transformer} and a peak learning rate of 0.002. We keep all of these hyperparameters consistent across each SSL model, only varying the batch size when memory limitations are encountered.

\subsubsection{JesusFilm ST: } \label{sec:a_st}
We collected the ST data from \url{jesusfilm.org}, which contains 2 hour long video dramas about the life of Jesus that are parallel in various languages. Each video contains multiple male and female speakers that appear throughout the drama. We downloaded the audio for Rajbanshi, Hijazi Arabic, and Lumun, while the ST labels are derived from the captions of the English audio. We process the data by splitting the 2-hour long videos sequentially, such that the first 70\% of the movie is used as the training set, the next 15\% is used as the development set, and the final 15\% is used as the test set. We use this method instead of random splitting to minimize the potential content and speaker overlap between the data splits. We manually clean the test set by filtering out segments with non-speech descriptions such as laughter or background music cues. For reference, the XEUS pre-training data contains 7 hours of Rajbanshi, 3 hours of Hijazi Arabic, and 0.5 hours of Lumun. 

To improve model convergence, we re-use the FLEURS models from Section \ref{sec:a_fleurs} and individually fine-tune them on ST for each language pair. We keep the fine-tuning parameters identical across all models, using a constant learning rate of 0.0001 with the Adam optimizer and a fixed batch size of 32. Inference is performed with a beam size of 10.

\subsubsection{Speech Resynthesis: }
Speech resynthesis experiments are performed on the VCTK dataset. For each SSL model we compare, we experiment with features extracted at 50\% and 75\% of the SSL model depth, which are standard settings used in discrete unit speech generation \cite{barrault2023seamlessm4t, maiti2024voxtlm}. For the 19-layer XEUS model, this means we test layers 10 and 14. For the 24-layer WavLM and w2v-BERT 2.0 v2 models, we trial layers 12 and 18. The features are then clustered at the frame-level via K-means to obtain the discrete units. We use a fixed value of $K=100$. We then train unit-to-speech HiFiGAN vocoders for each set of discrete units. Each vocoder is trained to generate 16 kHz speech for 150K steps. We use identical hyperparameters that correspond to the default VCTK settings in \url{https://github.com/kan-bayashi/ParallelWaveGAN} for each trial. We then select the best performing layer for each model based off of the MOSNet score of the resynthesized speech on the development set, and report its performance on the test set.

\subsubsection{SUPERB: } \label{sec:a_superb}
The SUPERB Benchmark tests SSL models on a diverse selection of tasks. The SSL model is frozen and a learned weighted sum of its layer-wise inputs are fed into the respective downstream probe fine-tuned for each task. To limit the hyperparameter search space, we only tune the learning rate and batch size. Otherwise, we use the same settings as the original benchmark \cite{superb}. For each task, the batch size is set to the maximum amount that can fit within a 40GB GPU. For the learning rate, we begin with the settings used by \citet{ChenWavLm} and conduct a sweep of those values multiplied by [0.25, 0.5, 1.0, 2.0, 4.0].

\begin{figure*}
    \centering
    \includegraphics[width=\textwidth]{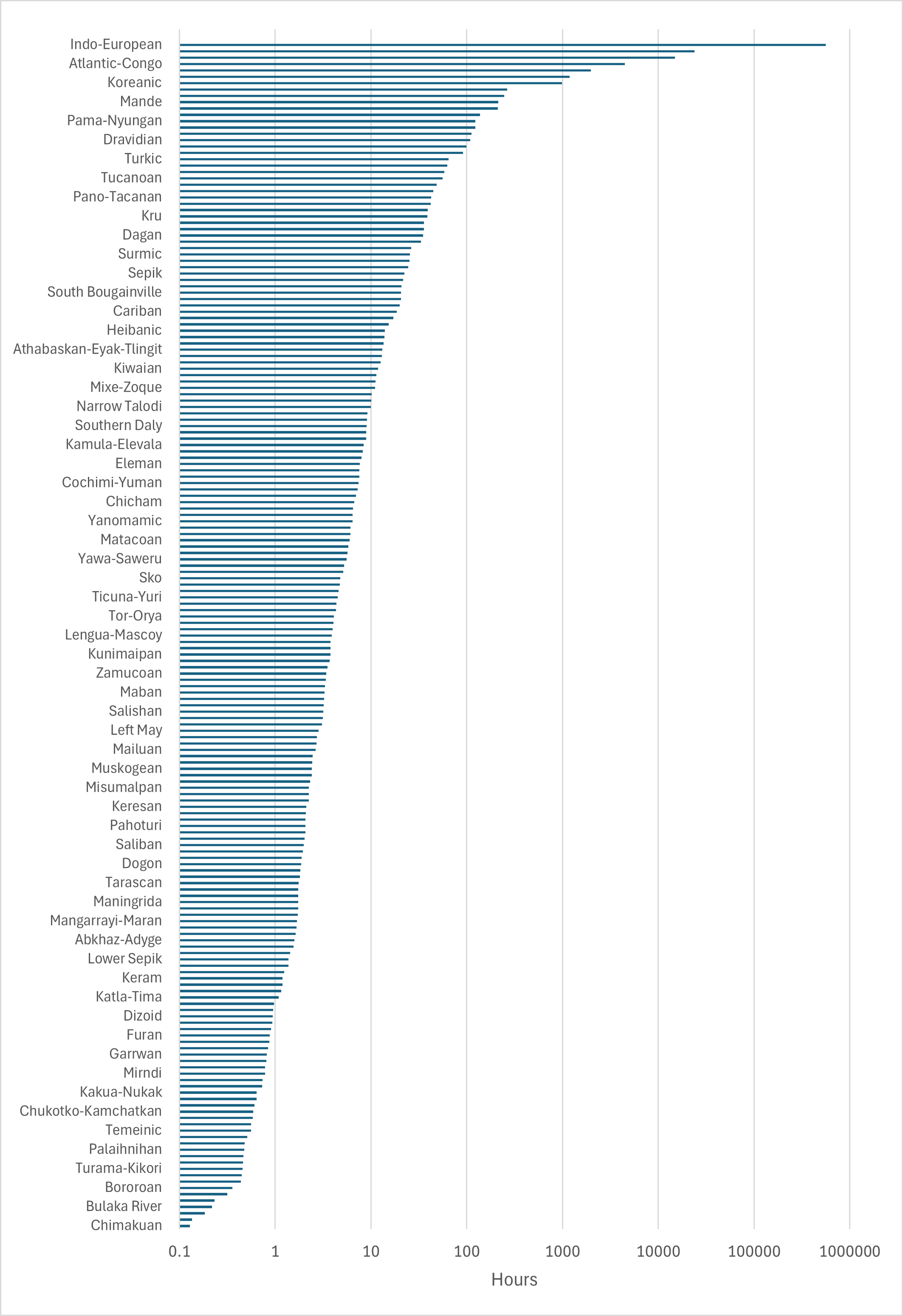}
    \caption{Distribution of data between the 189 language families in the XEUS pre-training data. We use Glottolog (\url{https://glottolog.org/}) to automatically map each ISO3 code to a language family.}
    \label{fig:families}
\end{figure*}

\end{document}